\documentclass[default,iicol]{sn-jnl}% Default with double column layout

\jyear{2021}%

\theoremstyle{thmstyleone}%
\theoremstyle{thmstyletwo}%

\theoremstyle{thmstylethree}%

\usepackage[ruled,vlined]{algorithm2e}
\usepackage{multirow}
\usepackage[caption=false,font=normalsize,labelfont=sf,textfont=sf]{subfig}
\begin{document}

\title[Article Title]{On the predictability in reversible steganography}


\author[1]{\fnm{Ching-Chun} \sur{Chang}}\email{ccchang@nii.ac.jp}

\author[2]{\fnm{Xu} \sur{Wang}}\email{xu.wang.phd@gmail.com}

\author[3]{\fnm{Sisheng} \sur{Chen}}\email{sisheng.chen.phd@gmail.com}

\author[4]{\fnm{Hitoshi} \sur{Kiya}}\email{kiya@tmu.ac.jp}

\author[1]{\fnm{Isao} \sur{Echizen}}\email{iechizen@nii.ac.jp}


\affil[1]{\orgname{National Institute of Informatics}, \orgaddress{\city{Tokyo}, \country{Japan}}}

\affil[2]{\orgdiv{Department of Information Engineering and Computer Science}, \orgname{Feng Chia University}, \orgaddress{\city{Taichung}, \country{Taiwan}}}

\affil[3]{\orgdiv{School of Big Data and Artificial Intelligence}, \orgname{Fujian Polytechnic Normal University}, \orgaddress{\city{Fuzhou}, \country{China}}}

\affil[4]{\orgdiv{Department of Computer Science}, \orgname{Tokyo Metropolitan University}, \orgaddress{\city{Tokyo}, \country{Japan}}}


\abstract{Artificial neural networks have advanced the frontiers of reversible steganography. The core strength of neural networks is the ability to render accurate predictions for a bewildering variety of data. Residual modulation is recognised as the most advanced reversible steganographic algorithm for digital images. The pivot of this algorithm is predictive analytics in which pixel intensities are predicted given some pixel-wise contextual information. This task can be perceived as a low-level vision problem and hence neural networks for addressing a similar class of problems can be deployed. On top of the prior art, this paper investigates predictability of pixel intensities based on supervised and unsupervised learning frameworks. Predictability analysis enables adaptive data embedding, which in turn leads to a better trade-off between capacity and imperceptibility. While conventional methods estimate predictability by the statistics of local image patterns, learning-based frameworks consider further the degree to which correct predictions can be made by a designated predictor. Not only should the image patterns be taken into account but also the predictor in use. Experimental results show that steganographic performance can be significantly improved by incorporating the learning-based predictability analysers into a reversible steganographic system.}




\maketitle

\section{Introduction}
In the age of automation and big-data analytics, the volume and variety of data from websites, social networks, electronic devices, and physical sensors are growing rapidly. The demand for maintaining data integrity and verifying data origin has soared with the data traffic growth. Malicious attacks such as data poisoning and adversarial perturbations impose cybersecurity threats to data-centric systems~\cite{2015_Perturb_Goodfellow, 2016_DeepFool, Poisoning17}. Steganography or invisible watermarking can serve as an authentication solution in data communications~\cite{668971, 959338, Abolfathi:2012aa}. It is the practice of hiding information into multimedia data and can be used to facilitate the management of authentication messages (e.g. cryptographic signatures, device fingerprints, and timestamps) by imperceptibly embedding such auxiliary information into the data. Reversibility prevents an accumulation of distortion over time, considering that digital media can be distributed over a series of electronic devices in which a new authentication message is embedded for each intermediate transmission. This reversible computing mechanism is particularly useful in integrity-sensitive applications such as creative production, forensic science, legal proceedings, military reconnaissance and medical diagnosis~\cite{Fridrich:2002aa, 2003_1196739, 2003_1227616, 2006_1608150, 2006_1608163, 2007_4291553, 2012_5957280, Feng:2012aa, 2016_RDH_Survey}.

In common with lossless compression, reversible computing depends primarily on data redundancy and predictability~\cite{1948_6773024}. Residual modulation is among the most advanced reversible steganographic methods for digital images~\cite{2007_4099409, 2008Fallahpour, 2014_6746082, Hwang:2016aa}. It embeds data by modulating pixels that can be accurately predicted. The remaining unpredictable pixels would also be modified to avoid pixel value overlap, though in this case distortion is caused without any capacity gain. The primary goal of steganographic algorithms is to minimise distortion (to ensure imperceptibility) subject to a given embedding rate (or capacity)~\cite{2002_1027818, Carli:2006aa, 2017_7393820}. Prediction accuracy has a decisive impact upon \emph{rate\textendash distortion} performance. A recent study has demonstrated that a multi-scale convolutional neural network (MS-CNN) can extract abundant image features based on multi-scale receptive fields, thereby achieving better accuracy than conventional predictors~\cite{Hu:2021aa}. Another study has established a connection between pixel-intensity prediction and low-level computer vision~\cite{Chang:2021aa}, and adopts a memory-persistent network (MemNet) originally proposed for super-resolution imaging and image denoising to the task of pixel-intensity prediction~\cite{8237748}. 

While it is possible to improve prediction accuracy by building a sophisticated predictive model, non-trivial prediction errors may still occur when making predictions for certain rare and complex patterns. If predictability is quantifiable, we can select the most predictable pixels for carrying the payload, thereby minimising unnecessary distortion. Predictability is content-dependent: in general, it is high for smooth regions and low for complex regions. To measure the smoothness of a local image patch, we may formulate a function that analyses pixel correlations and estimates pattern complexity, often by calculating the variance of neighbouring pixels~\cite{2009_4811982, 2011_5762603, Peng:2012aa, Cao:2019aa}. However, this kind of prescriptive statistical method appears to fall short of capturing true predictability (or predictive uncertainty), defined as the degree to which a correct prediction can be made by a given predictor. Predictability is not only associated with image patterns but also with the predictor used. Hence, we propose to analyse directly which types of patterns a given predictor has the highest probability of making accurate predictions.

In this paper, we propose both \emph{supervised} and \emph{unsupervised} learning for training predictability analysers by which each pixel in an image is assigned with a predictability score. Learning-based predictability analysers can leverage statistics of image patterns and simultaneously adapt to a specific predictor. For supervised learning, we construct a ground-truth dataset by representing predictability as quantised residuals and train a neural network to segment predictable and unpredictable regions. For unsupervised learning, we formulate a dual-headed neural network trained with an ad hoc loss function to make predictions and estimate predictability simultaneously. We carry out an ablation study to clarify the contribution of the learning-based predictability analysers to the steganographic system and compare their performance with that of a representative statistical analyser.

The remainder of this paper is organised as follows. Section~\ref{sec:back} provides the background regarding the recent development of reversible steganography. Section~\ref{sec:method} presents the proposed predictability analysis methods. Section~\ref{sec:exp} evaluates the effects of the proposed methods in comparison to the benchmarks. Concluding remarks are given in Section~\ref{sec:con}.
%%% ----- END OF SECTION ----- %%%
%%% ----- END OF SECTION ----- %%%
%%% ----- END OF SECTION ----- %%%

\section{Background}
\label{sec:back}
Advances in deep learning have led to significant breakthroughs in many branches of science~\cite{LeCun:2015aa}. Steganography with deep learning has emerged as a promising research paradigm in which deep neural networks are used as a part or a whole of the steganographic system~\cite{NIPS2017_838e8afb, NIPS2017_6791, 2017_8017430, Zhu:2018aa, Volkhonskiy:2019aa, Wengrowski_2019_CVPR, Tancik:2020aa, Luo:2020aa}. Current reversible steganographic methods with deep learning can be categorised into \emph{end-to-end} and \emph{modular} pipelines. Typically, the end-to-end pipeline trains a pair of encoder and decoder networks jointly to simulate a whole reversible steganographic system~\cite{Jung:2019aa, Duan:2019aa, Lu_2021_CVPR}. These networks can learn to embed/extract confidential messages into/from carrier signals automatically. While such a monolithic end-to-end system usually offers high steganographic capacity, perfect reversibility cannot be guaranteed. Deep learning is powerful for handling the nonlinear nature of the real world; however, reversible computing is more of a mechanical process in which procedures have to be conducted in accordance with rigorous algorithms. One way to overcome the problem of imperfect reversibility is through modularisation. The modular pipeline deploys neural networks in a particular module (typically for analysing data distribution or data redundancy) while handling reversibility through established reversible steganographic algorithms~\cite{2020_9245471}.

Residual modulation has a modular pipeline that consists of a \emph{predictive analytics} module and a \emph{reversible coding} module. The former predicts pixel intensity based on some pixel-wise contextual information, whereas the latter encodes/decodes a message bit-stream into/from residuals (i.e. prediction errors) and is responsible for reversibility. A more accurate predictor leads to a better rate\textendash distortion trade-off (i.e. the balance between capacity and imperceptibility). There are basically two types of approaches to programming predictive algorithms. The rule-based approach formulates a predictive model with handcrafted or fixed parameters~\cite{2007_4099409, 2008Fallahpour, 2014_6746082, Hwang:2016aa}. This type of model typically has low computational complexity, so it can be implemented with a minimal number of parameters and deployed in real-time applications. The learning-based approach finds optimal parameters for a predictive model by feeding it with vast amounts of data (e.g. MS-CNN~\cite{Hu:2021aa} and MemNet~\cite{Chang:2021aa}). A neural network model usually has millions of learnable parameters to deal with a bewildering range of input images and can learn to extract contributory features automatically for making precise predictions~\cite{NIPS2012_AlexNet, Simonyan15, 2015_7298594}.

One way to improve prediction accuracy is to construct a sophisticated neural network model. Another way is to adaptively filter out some unpredictable regions. As a consequnece, rate\textendash distortion performance would be improved because the remaining pixels are more likely to be accurately predicted. As a side effect, however, some potential carrier pixels may also be filtered out, thereby decreasing the maximum capacity. To remedy this problem, instead of filtering out pixels, we embed the payload in from the most to the least predictable pixels.

%%% ----- FIGURE ----- %%%
\begin{figure*}[t]
\centerline{\includegraphics[width=1.98\columnwidth]{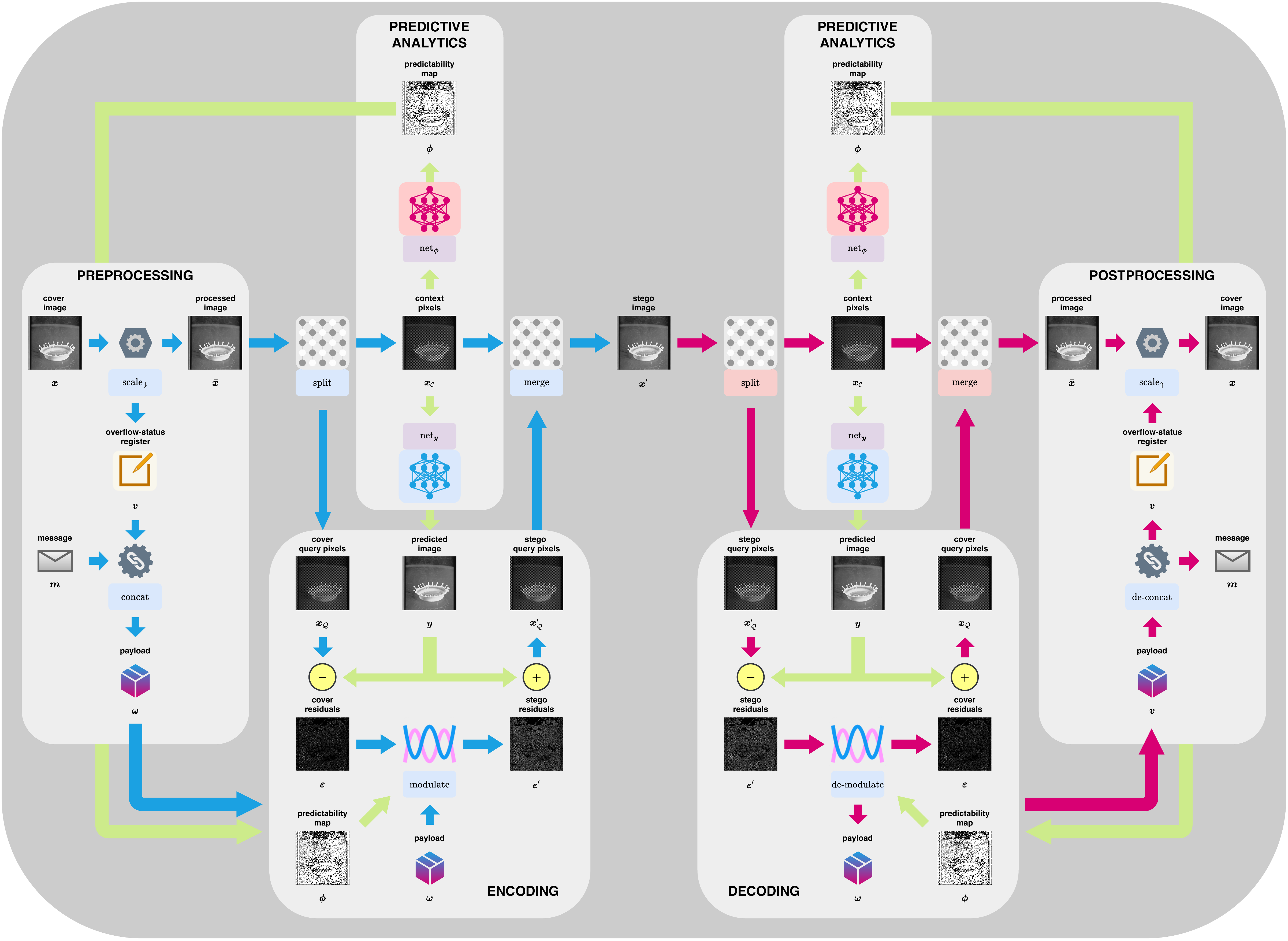}}
\caption{Schematic workflow of reversible steganography with adaptive residual modulation.}
\label{fig:scheme}
\end{figure*}
%%% ----- END OF FIGURE ----- %%%

%%% ----- TABLE ----- %%%
\begin{table}[t]
\caption{List of variables and operations} % title of Table
\centering % used for centering table
\begin{tabular}{l l} % centered columns (4 columns)
\hline\hline %inserts double horizontal lines
\textbf{var. \& op.} & \textbf{definition}\\ [0.5ex] % inserts table
%heading
\hline % inserts single horizontal line
$\boldsymbol{x}$ 								& cover image \\
$\boldsymbol{x}^{\prime}$ 						& stego image \\
$\bar{\boldsymbol{x}}$ 				& processed image \\
$\boldsymbol{x}_{\mathcal{C}}$ 					& context pixels \\
$\boldsymbol{x}_{\mathcal{Q}}$ 					& query pixels \\
$\boldsymbol{y}_{\mathcal{C}}$ 					& predicted context pixels \\
$\boldsymbol{y}_{\mathcal{Q}}$ 					& predicted query pixels \\
$\boldsymbol{\varepsilon}$ 						& prediction residuals\\
$\boldsymbol{m}$ 								& message \\
$\boldsymbol{v}$ 								& overflow-status register \\
$\boldsymbol{\omega}$ 							& payload \\
$\boldsymbol{\phi}$ 								& predictability map\\
$\boldsymbol{\sigma}^2$ 								& uncertainty/variance map\\
$\alpha$ 							& stego-channel parameter\\
$\operatorname{net}_{\phi}$ 			& neural network for predictability estimation\\
$\operatorname{net}_{y}$ 			& neural network for intensity prediction\\
$\operatorname{scale}_{\Downarrow}$	& downscaling of possibly overflowing pixels\\
$\operatorname{scale}_{\Uparrow}$ 	& upscaling of possibly overflowing pixels\\
$\operatorname{split}$ 				& context-query splitting \\
$\operatorname{merge}$ 				& context-query merger\\
$\operatorname{select}$ 				& adaptive selection of predictable pixels \\
$\operatorname{replace}$ 			& replacement of predictable pixels\\
$\operatorname{modulate}$ 			& residual modulation (payload embedding)\\
$\operatorname{de-modulate}$ 		& residual de-modulation (payload extraction)\\
$\operatorname{concat}$ 				& concatenation of bit-streams\\
$\operatorname{de-concat}$ 			& de-concatenation of bit-streams\\
[1ex] % [1ex] adds vertical space
\hline %inserts single line
\end{tabular}
\label{tab:def} % is used to refer this table in the text
\end{table}
%%% ----- END OF TABLE ----- %%%

%%% ----- SECTION ----- %%%
%%% ----- SECTION ----- %%%
%%% ----- SECTION ----- %%%
\section{Methodology}
\label{sec:method}
We start with a system overview that outlines the encoding/decoding procedures, followed by algorithmic details of the steganographic mechanisms. Then, we present a neural network model for pixel intensity prediction, which leads to the notion of predictability analysis. Both supervised and unsupervised learning frameworks are introduced for analysing predictability, thereby creating an adaptive route for data embedding. Furthermore, a method for analysing pattern complexity based on local variance is established as a benchmark to which the proposed predictability analysis methods are compared.

%%% ----- SUBSECTION ----- %%%
%%% ----- SUBSECTION ----- %%%
\subsection{System Overview}
The objective of reversible steganography is to modify a \emph{cover} image $\boldsymbol{x}$ in an imperceptible and reversible manner to embed a message $\boldsymbol{m}$, resulting in a \emph{stego} image $\boldsymbol{x}^{\prime}$. A reversible steganographic system can be divided into an encoding phase (message embedding) and a decoding phase (message extraction and image restoration), as outlined in Figure~\ref{fig:scheme}. The encoding is operated as follows. To begin with, the image is pre-processed to prevent pixel intensity overflow. This pre-processing is performed by scaling down \emph{near-saturated} pixels (i.e. pixels with an intensity value around the maximum or minimum). An overflow-status register $\boldsymbol{v}$ is used to flag the processed pixels in order to distinguish them from unprocessed ones with the same values. It is required for perfect reversibility, and thus should be concatenated with the intended message as a part of the payload $\boldsymbol{\omega}$. The image is split into the \emph{context} and \emph{query} pixel sets, denoted respectively by $\boldsymbol{x}_{\mathcal{C}}$ and $\boldsymbol{x}_{\mathcal{Q}}$, via a pre-defined sampling lattice. Predictions of query pixels (denoted by $\boldsymbol{y}$) along with a predictability map (denoted by $\boldsymbol{\phi}$) are derived from the context pixels by applying either two separate neural networks (i.e. supervised learning framework) or a single neural network (i.e. unsupervised learning framework). The residuals between the pixels and their predicted counterpart, as denoted by $\boldsymbol{\varepsilon}$ are computed. The payload is embedded by modulating the residuals. The route of madulation is determined by the predictability map. In other words, the payload is embedded in from the most to the least predictable pixels. The query pixels are modified by adding the modulated residuals to the predicted intensities. The context and query pixels are then merged together to obtain a stego image. The decoding is operated in a similar way. It begins by splitting the stego image into the context and query sets. Since the context set remains unmodified during the encoding phase, the same predictions, as well as an identical predictability map, can be derived. Then, the residuals are calculated and de-modulated, the payload is extracted, and finally, the query pixels are restored. Post-processing is required to segment the payload into the message and overflow-status register. The latter is then used to scale up those near-saturated pixels to their original intensities, yielding a distortion-free image. A list of variables and operations is provided in Table~\ref{tab:def}, and a concise summary of encoding and decoding phases is shown in Algorithms~\ref{alg:enc} and~\ref{alg:dec}.

%%% ----- ALGORITHM ----- %%%
\begin{algorithm}[t!]
\SetAlgoLined
\KwIn{$\boldsymbol{x}$ and $\boldsymbol{m}$}
\KwOut{$\boldsymbol{x}^{\prime}$}

\tcp{pre-processing}
\begin{itemize}
\item $[\bar{\boldsymbol{x}}, \boldsymbol{v}] = \operatorname{scale}_{\Downarrow}(x \mid \alpha)$
\item $\boldsymbol{\omega} = \operatorname{concat}(\boldsymbol{m}, \boldsymbol{v})$
\end{itemize}

\tcp{predictive analytics}
\begin{itemize}
\item $[\boldsymbol{x}_{\mathcal{C}}, \boldsymbol{x}_{\mathcal{Q}}] = \operatorname{split}(\bar{\boldsymbol{x}})$
\item $[\boldsymbol{y}_{\mathcal{C}}, \boldsymbol{y}_{\mathcal{Q}}] = \operatorname{net}_{\boldsymbol{y}}( [\boldsymbol{x}_{\mathcal{C}}, \boldsymbol{0}] )$
\item $\boldsymbol{\phi} = \operatorname{net}_{\boldsymbol{\phi}}( [\boldsymbol{x}_{\mathcal{C}}, \boldsymbol{0}] )$
\end{itemize}

\tcp{encoding}
\begin{itemize}
%\item $[\tilde{\boldsymbol{x}}_{\mathcal{Q}}, \tilde{\boldsymbol{y}}] = \operatorname{select}(\boldsymbol{x}_{\mathcal{Q}}, \boldsymbol{y} \mid \boldsymbol{\phi})$
\item $\boldsymbol{\varepsilon} = \boldsymbol{x}_{\mathcal{Q}} - \boldsymbol{y}_{\mathcal{Q}}$
\item $\boldsymbol{\varepsilon}^{\prime} = \operatorname{modulate}(\boldsymbol{\varepsilon}, \boldsymbol{\omega} \mid \boldsymbol{\phi}, \alpha)$
\item $\boldsymbol{x}_{\mathcal{Q}}^{\prime} = \boldsymbol{y}_{\mathcal{Q}} + \boldsymbol{\varepsilon}^{\prime}$
%\item $\boldsymbol{x}_{\mathcal{Q}}^{\prime} = \operatorname{replace}(\boldsymbol{x}_{\mathcal{Q}}, \tilde{\boldsymbol{x}}_{\mathcal{Q}}^{\prime} \mid \boldsymbol{\phi})$
\item $\boldsymbol{x}^{\prime} = \operatorname{merge}(\boldsymbol{x}_{\mathcal{C}}, \boldsymbol{x}_{\mathcal{Q}}^{\prime})$
\end{itemize}

\caption{Encoding}
\label{alg:enc}
\end{algorithm}
%%% ----- END OF ALGORITHM ----- %%%
%%% ----- ALGORITHM ----- %%%
\begin{algorithm}[t!]
\SetAlgoLined
\KwIn{$\boldsymbol{x}^{\prime}$}
\KwOut{$\boldsymbol{x}$ and $\boldsymbol{m}$}

\tcp{predictive anlaytics}
\begin{itemize}
\item $[\boldsymbol{x}_{\mathcal{C}}, \boldsymbol{x}_{\mathcal{Q}}^{\prime}] = \operatorname{split}(\boldsymbol{x}^{\prime})$
\item $[\boldsymbol{y}_{\mathcal{C}}, \boldsymbol{y}_{\mathcal{Q}}] = \operatorname{net}_{\boldsymbol{y}}( [\boldsymbol{x}_{\mathcal{C}}, \boldsymbol{0}] )$
\item $\boldsymbol{\phi} = \operatorname{net}_{\boldsymbol{\phi}}( [\boldsymbol{x}_{\mathcal{C}}, \boldsymbol{0}] )$
\end{itemize}

\tcp{decoding}
\begin{itemize}
%\item $[\tilde{\boldsymbol{x}}_{\mathcal{Q}}^{\prime}, \tilde{\boldsymbol{y}}] = \operatorname{select}(\boldsymbol{x}_{\mathcal{Q}}^{\prime}, \boldsymbol{y} \mid \boldsymbol{\phi})$
\item $\boldsymbol{\varepsilon}^{\prime} = \boldsymbol{x}_{\mathcal{Q}}^{\prime} - \boldsymbol{y}_{\mathcal{Q}}$
\item $[\boldsymbol{\varepsilon}, \boldsymbol{\omega}] = \operatorname{de-modulate}(\boldsymbol{\varepsilon}^{\prime} \mid \boldsymbol{\phi}, \alpha)$
\item $\boldsymbol{x}_{\mathcal{Q}} = \boldsymbol{y}_{\mathcal{Q}} + \boldsymbol{\varepsilon}$
%\item $\boldsymbol{x}_{\mathcal{Q}} = \operatorname{replace}(\boldsymbol{x}_{\mathcal{Q}}^{\prime}, \tilde{\boldsymbol{x}}_{\mathcal{Q}} \mid \boldsymbol{\phi})$
\item $\bar{\boldsymbol{x}} = \operatorname{merge}(\boldsymbol{x}_{\mathcal{C}}, \boldsymbol{x}_{\mathcal{Q}})$
\end{itemize}

\tcp{post-processing}
\begin{itemize}
\item $[\boldsymbol{m}, \boldsymbol{v}] = \operatorname{de-concat}(\boldsymbol{\omega})$
\item $\boldsymbol{x} = \operatorname{scale}_{\Uparrow}(\bar{\boldsymbol{x}} , \boldsymbol{v}  \mid \alpha)$
\end{itemize}

\caption{Decoding}
\label{alg:dec}
\end{algorithm}
%%% ----- END OF ALGORITHM ----- %%%

\subsection{Algorithmic Details}
Let us denote by $\alpha$ a \emph{stego-channel} parameter that regulates how a residual is modulated, where $\alpha \in \mathbb{Z}^{+}$. It acts as a threshold value between the \emph{carrier} and \emph{non-carrier} residuals, such that
\begin{equation}
\varepsilon \in 
\begin{cases}
\textrm{carrier} &\textrm{if } \operatorname{abs}(\varepsilon) < \alpha ,  \\
\textrm{non-carrier} &\textrm{otherwise},
\end{cases}
\end{equation}
where $\operatorname{abs}(\varepsilon)$ denotes the absolute magnitude of a residual. According to the law of error, the frequency of a residual can be expressed as an exponential function of its magnitude and therefore follows a zero-mean Laplacian distribution~\cite{Wilson:1923aa}. In other words, the residuals with magnitude $0$ should occur most frequently, followed by $\pm{1}$, $\pm{2}$, and so forth. To embed a message, we modulate a residual symmetrically from magnitude $0$ to $\pm{1}$ or keep it unchanged, giving three different states, thereby enabling one ternary digit ($\log_2 3 \approx 1.58$ bits of information) to be represented. To circumvent the inherent problem with numerical base conversion, we embed one or two bits with a probability of $0.5$ each in the following way:
\begin{equation}
\{\varepsilon^{\prime} \mid \varepsilon=0\} =
\begin{cases}
0 &\textrm{if $\omega_t = 0$} ,\\
-1 &\textrm{if $[\omega_t, \omega_{t+1}] = [1,0]$} ,\\
+1 &\textrm{if $[\omega_t, \omega_{t+1}] = [1,1]$} ,
\end{cases}
\end{equation}
where $\omega_t$ denotes the $t$\textsuperscript{th} payload bit. This compromise solution enables the embedding of $1.5$ bits on average. The absolute magnitude of each carrier residual other than those with magnitude $0$ is either kept unchanged or increased by $1$ to embed one bit of information; that is,
\begin{equation}
\{\varepsilon^{\prime} \mid 0 < \operatorname{abs}(\varepsilon) < \alpha \} =
\begin{cases}
2\varepsilon &\textrm{if $\omega_t = 0$} ,\\
2\varepsilon + \operatorname{sgn}(\varepsilon)\cdot 1 &\textrm{if $\omega_t = 1$} ,
\end{cases}
\end{equation}
where $\operatorname{sgn}(\varepsilon)$ denotes the sign of a residual (either positive or negative). For each non-carrier residual, we increase its absolute magnitude by $\alpha$ to separate it from other modulated residuals; that is,
\begin{equation}
\{\varepsilon^{\prime} \mid \operatorname{abs}(\varepsilon) \geq \alpha \} = \varepsilon + \operatorname{sgn}(\varepsilon)\cdot \alpha .
\end{equation}
The modulation in the residual domain produces a pro rata increase or decrease in the spatial domain. Decoding is simply an inverse mapping. Let $x_{i,j}$ denote a pixel at co-ordinates $(i,j)$. To prevent overflow, we scale down near-saturated pixels (determined by $\alpha$) in advance:
\begin{equation}
x_{i,j} = 
\begin{cases}
x_{i,j} + \alpha &\textrm{if } x_{i,j} \in [x_{\textrm{min}}, x_{\textrm{min}} + \alpha - 1],\\
x_{i,j} - \alpha &\textrm{if } x_{i,j} \in [x_{\textrm{max}} - \alpha + 1, x_{\textrm{max}}],\\
\end{cases}
\end{equation}
where $x_{\textrm{min}}$ and $x_{\textrm{max}}$ denote the minimum and maximum of all possible pixel values, respectively (e.g. values $0$ and $255$ for $8$-bit colour depth). The downscaling causes collisions between the scaled and unscaled pixel values. Such collisions are recorded by setting flags to $1$ (true) for the scaled pixels and $0$ (false) for the unscaled pixels. These flag bits are stored in the overflow-status register $\boldsymbol{v}$.

%%% ----- FIGURE ----- %%%
\begin{figure}[t]
\centerline{\includegraphics[width=0.5\columnwidth]{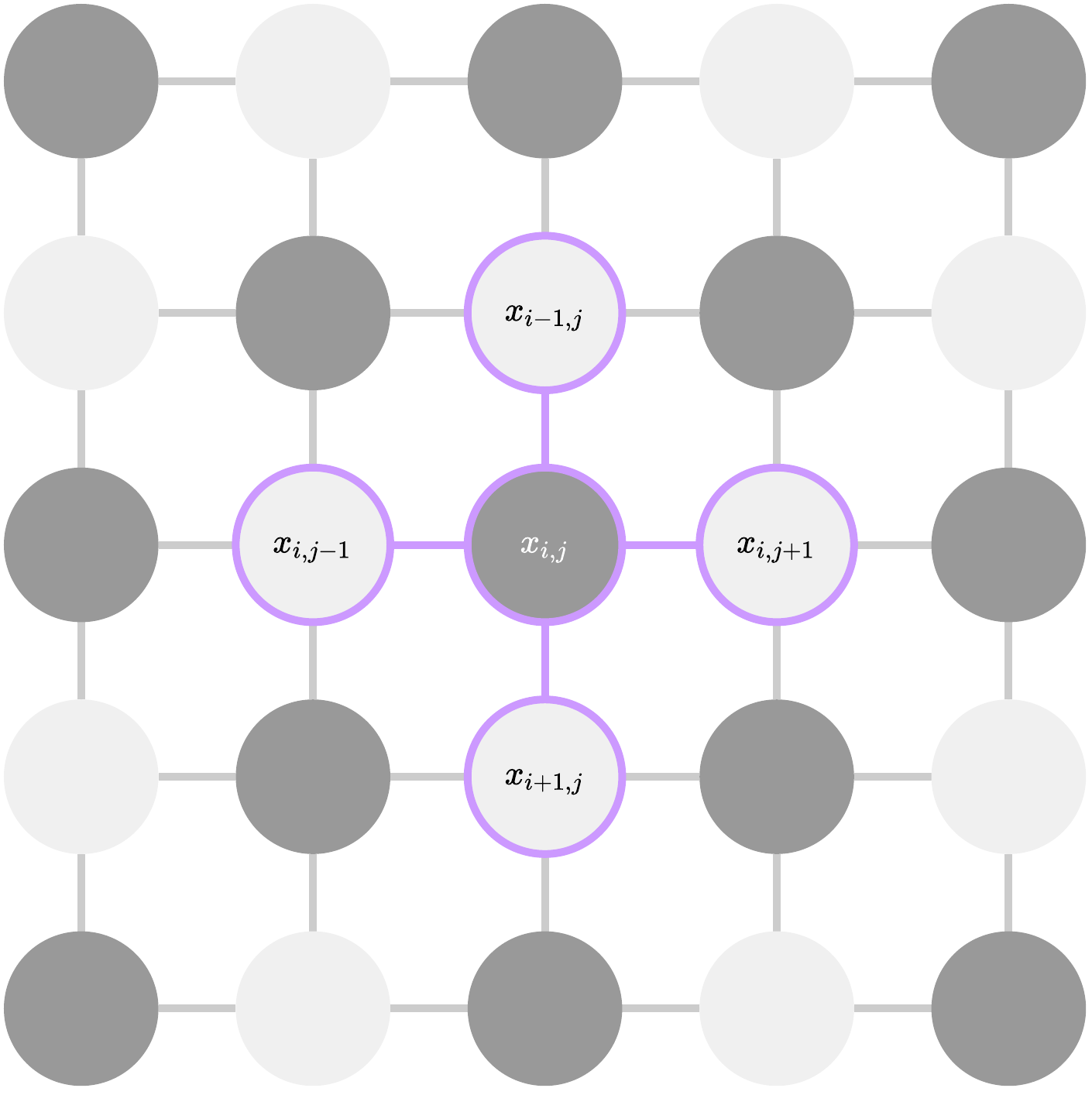}}
\caption{Chequerboard-like lattice derived from four-pixel connectivity for context/query splitting.}
\label{fig:lattice}
\end{figure}
%%% ----- END OF FIGURE ----- %%%

%%% ----- SUBSECTION ----- %%%
%%% ----- SUBSECTION ----- %%%
\subsection{Predictive Analytics}
\begin{figure}[t]
\centerline{\includegraphics[width=0.98\columnwidth]{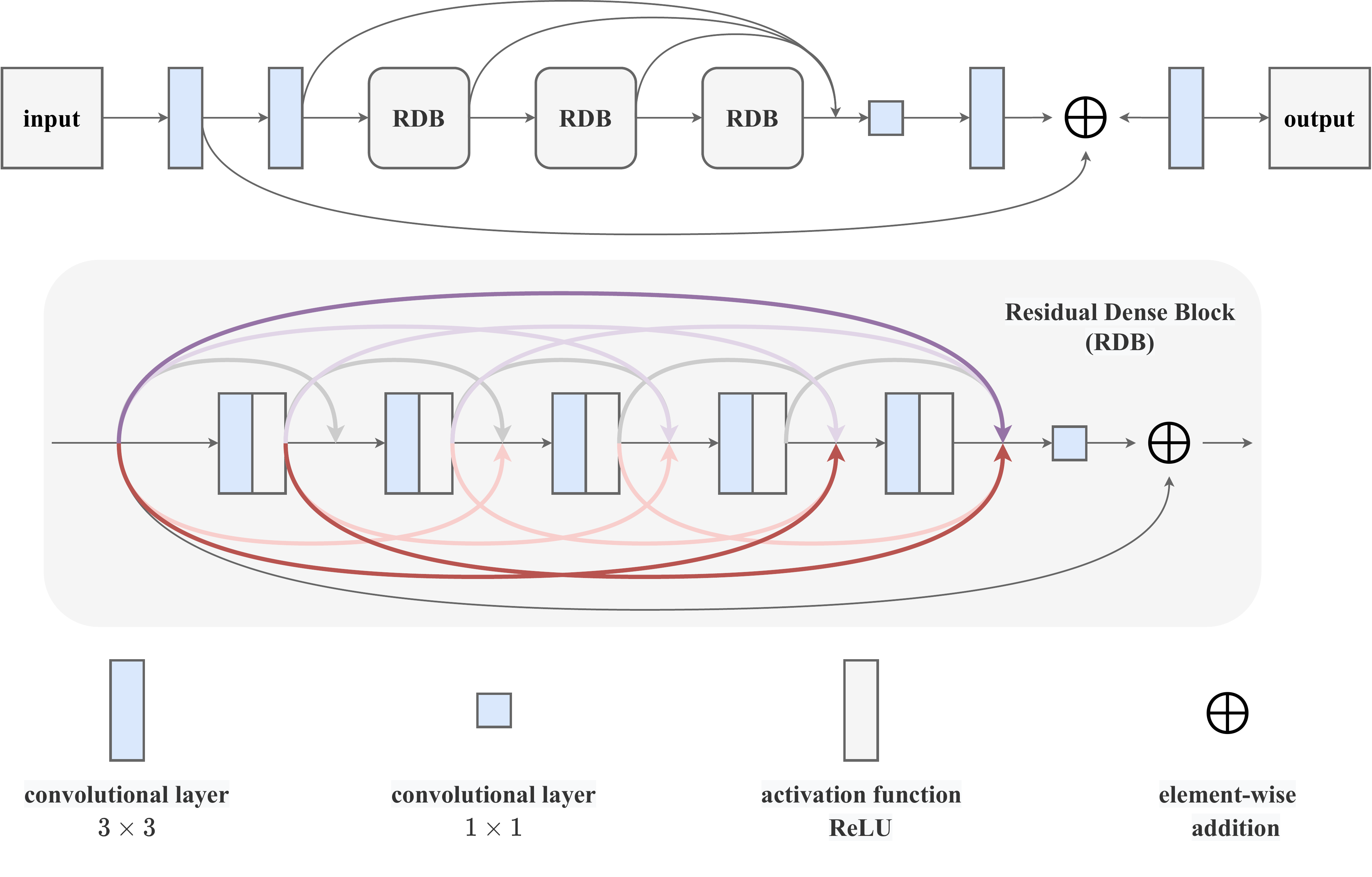}}
\caption{Architecture of residual dense network for intensity prediction.}
\label{fig:RDN}
\end{figure}

Intensity prediction aims to extrapolate the intensities of query pixels from the observed context pixels. A common way to sample the context and query pixels is by using a chequerboard-like lattice, as illustrated in Figure~\ref{fig:lattice}. The lattice forms a uniform sampling pattern such that each query pixel is surrounded by four context pixels connected horizontally and vertically. This particular prediction task can be viewed as a low-level vision problem that concerns spatial correlations as well as low-level image features (e.g. colour, texture, and shape). This perception enables intensity prediction to be handled with neural network models designed for solving a similar class of problems such as super-resolution reconstruction, missing-data restoration, and noise reduction. The input to the model is an image with original intensities at the context pixels and zeros at the query pixels. The target is the original image. The residual dense network (RDN) is considered state of the art for such low-level vision problems~\cite{2018_8578360}. The RDN model is characterised by its network-in-network architecture comprising a tangled labyrinth of residual and dense connections, as illustrated in Figure~\ref{fig:RDN}. It has a hierarchical structure that extracts image features at both local and global levels. At the local level, a number of $3 \times 3$ convolutional layers are densely connected with a non-linear ReLU activation function applied after each layer~\cite{2011_ReLU}. The outputs from previous layers are then integrated by a $1 \times 1$ convolutional layer that fuses multiple features by their saliencies~\cite{Lin:2014aa}. A skip connection between the input and output of this residual dense block (RDB) is established to mitigate the problem of vanishing gradients~\cite{2016_7780459}. At the global level, the ouputs from several RDBs are fused by a $1 \times 1$ convolutional layer with a skip connection between a shallow layer and a deep layer~\cite{8964437}. In our implementation, the number of RDBs is specified as $3$ and the number of convolutional layers in each RDB is configured to $5$. The RDN is trained to minimise the $\ell_1$ norm between the predicted outputs and the targets via the back-propagation algorithm~\cite{Rumelhart:1986ab}.

%%% ----- FIGURE ----- %%%
%%% ----- END OF FIGURE ----- %%%

Although neural networks are capable of making accurate predictions on average, the predicted intensities of particular pixels can be heavily skewed. Identifying and filtering out the unpredictable pixels should further increase the overall accuracy. These singular pixels often appear in some highly textured areas. While it is possible to capture this content-dependent nature by using a pattern analysis algorithm, we argue that the identified complex regions are not completely equivalent to the desired unpredictable regions. We thus propose estimating predictability directly. In the following, we introduce both supervised and unsupervised learning for predictability analysis. From a certain perspective, predictability analysis or uncertainty quantification is an essential element towards self-aware intelligent machinery. The concept of \emph{self-awareness}, albeit elusive and indefinite, can be understood as the ability to know one's own strengths and limitations. While there is no indication that contemporary artificial intelligence is anywhere close to engendering self-awareness, a rudimentary kind of self-aware machine should be able to realise whether or not it can handle a given query: the machine knows what it knows and does not know. With such self-awareness, the machine can be given a reject option and be allowed to abstain from making a prediction when there is a large amount of uncertainty, thereby leading to robust and reliable decision-making~\cite{827457, pmlr-v97-geifman19a, pmlr-v97-thulasidasan19a, Kompa:2021aa}.

%%% ----- END OF SECTION ----- %%%
%%% ----- END OF SECTION ----- %%%
%%% ----- END OF SECTION ----- %%%

%%% ----- FIGURE ----- %%%
\begin{figure}[t]
\centerline{\includegraphics[width=0.85\columnwidth]{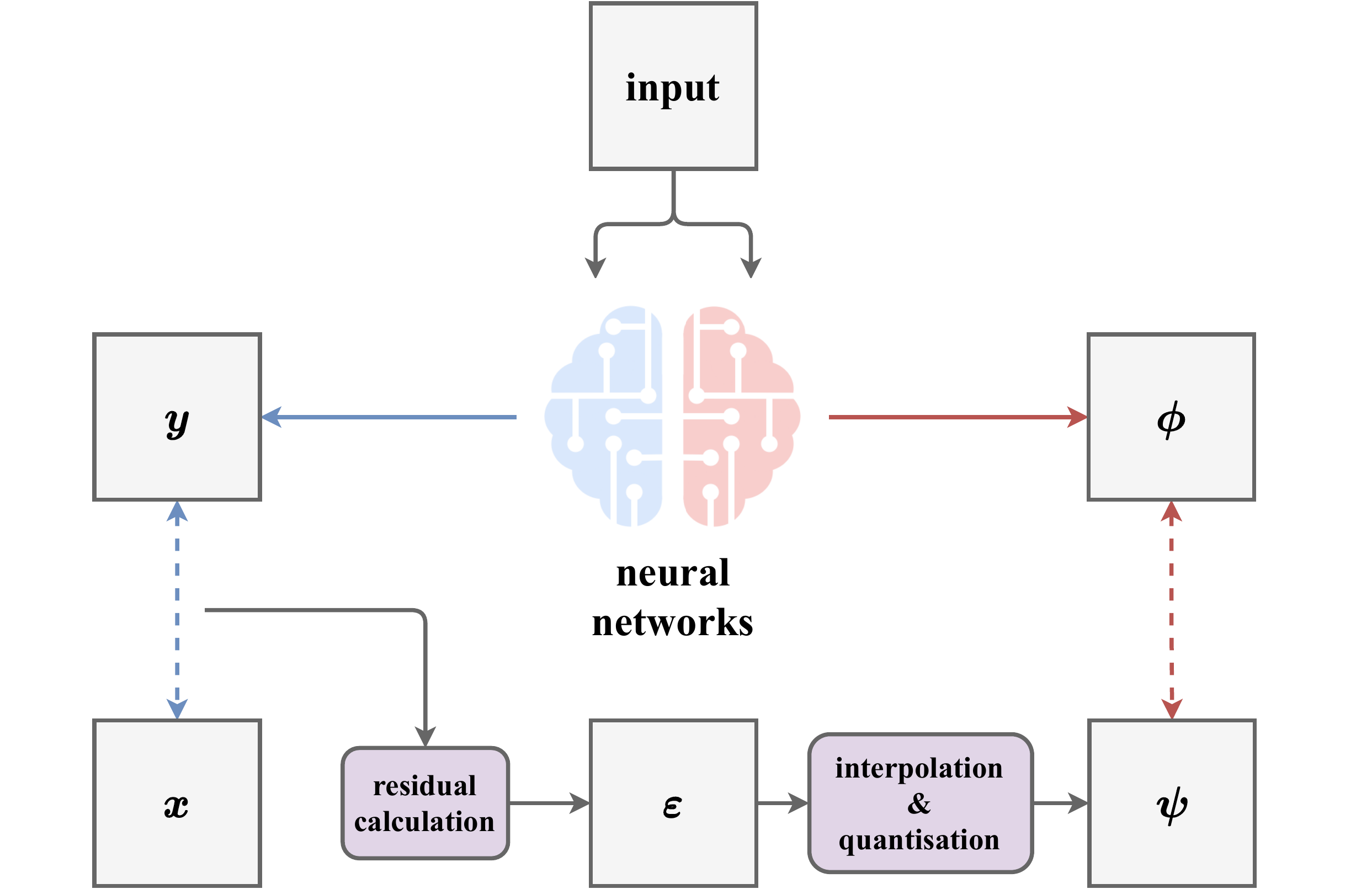}}
\caption{Supervised learning for intensity prediction and predictability analysis with ground-truth generation.}
\label{fig:sup_learn}
\end{figure}
%%% ----- END OF FIGURE ----- %%%

\subsection{Predictability Analysis with Supervised Learning}
Let us consider a simplified case in which each pixel is classified as either predictable or unpredictable, denoted by $1$ and $0$ respectively. This is a pixel-level binary classification problem, which can also be viewed as a special kind of image segmentation problem. This means we can employ a neural network model originally developed for image segmentation to deal with predictability analysis. The remaining question is how to construct the ground-truth targets to which the model is trained to fit. To enable supervised learning, we create the ground-truth targets by quantising the prediction residuals to $1$ if their magnitudes are less than a given threshold and to $0$ otherwise. In practice, we set this threshold to $\alpha$ to link the predictable pixels with the carrier pixels. The task is therefore identifying potential carrier pixels. Since residuals can only be computed for query pixels, we approximate the residuals for context pixels by linear interpolation\textemdash that is, the mean of four neighbours. A diagram of the ground-truth generation process is shown in Figure~\ref{fig:sup_learn}. In the implementation, we deploy the classic U-shaped network (U-Net) widely used for image segmentation~\cite{2015_UNet}. This model comprises a pair of encoder and decoder networks with skip connections between mirrored layers. The image passes from top to bottom through consecutive downscaling convolutions and is then updated through a bottom-up pass with a succession of upscaling convolutions. The model learns to assemble a precise output on the basis of multi-resolution features. A sigmoid activation function is placed at the end of the model to bound the output values between $0$ and $1$. The U-Net is trained to minimise the cross-entropy loss function:
\begin{equation}
\mathcal{L}_{\boldsymbol{\phi}} = -\sum_i \sum_j \psi_{i,j} \log(\phi_{i,j}) + (1-\psi_{i,j}) \log(1-\phi_{i,j}),
\end{equation}
where $\psi_{i,j}$ is the ground truth for $\phi_{i,j}$.

%%% ----- FIGURE ----- %%%
\begin{figure}[t]
\centerline{\includegraphics[width=0.99\columnwidth]{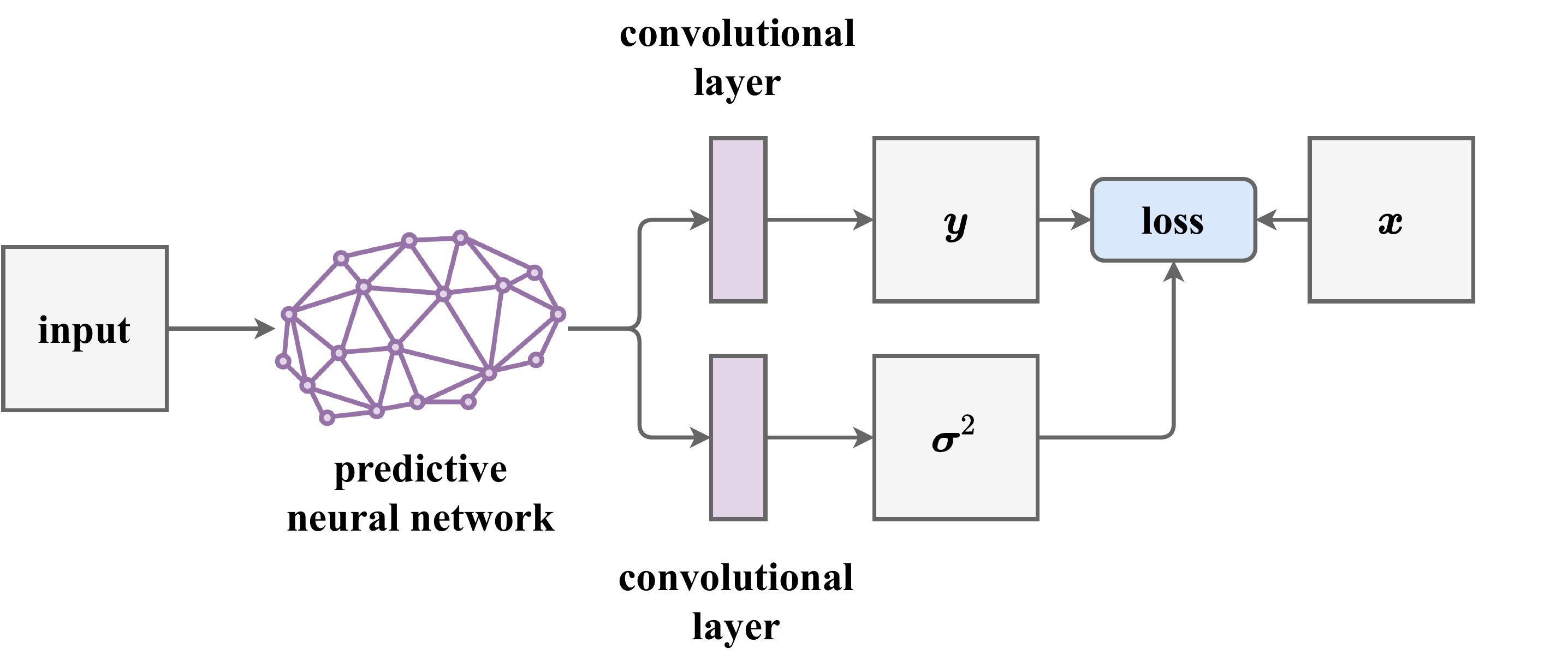}}
\caption{Unsupervised learning for uncertainty quantification with dual-headed neural network.}
\label{fig:unsup_learn}
\end{figure}
%%% ----- END OF FIGURE ----- %%%

\subsection{Predictability Analysis with Unsupervised Learning}
Learning to generate predictability maps in a supervised fashion, albeit feasible, involves extensive computational time and memory for constructing a ground-truth dataset and additional storage space for the predictability analysis model. It is possible to analyse predictability in a partially unsupervised or self-supervised manner. Rather than training an extra model, we add an extra output branch to the existing predictive model for estimating predictability. Let us denote by $\boldsymbol{x}$ the ground-truth image, $\boldsymbol{y}$ the image predicted by the pre-trained RDN, and $\boldsymbol{\sigma}^2$ the predictive uncertainty map to be learnt. A smaller value of $\sigma_{i,j}^2$ indicates a higher confidence in predicting the corresponding pixel. We append an additional trainable convolutional layer prior to the last layer of the pre-trained neural network. This dual-branch neural network, as illustrated in Figure~\ref{fig:unsup_learn}, is trained to minimise the following loss function:
\begin{equation}
\mathcal{L}_{\boldsymbol{\phi}} = \mathcal{C}(\boldsymbol{x}, \boldsymbol{y}\mid \boldsymbol{\phi}) + \lambda \mathcal{R}(\boldsymbol{\phi}) ,
\end{equation}
where $\mathcal{C}$ is the cost function and $\mathcal{R}$ is the regularisation term weighted by $\lambda$. The cost function is defined by
\begin{equation}
\mathcal{C}(\boldsymbol{x}, \boldsymbol{y}\mid \boldsymbol{\phi}) = \sum_i \sum_j \frac{ \lvert x_{i,j} - y_{i,j} \rvert}{\sigma_{i,j}^2}.
\end{equation}
This weighted distance term discourages the model from causing high regression residuals with low uncertainty and attenuates loss when having high uncertainty. The regularisation term is used to penalise the model for being uncertain, thereby encouraging it to reveal more predictable regions, as defined by
\begin{equation}
\mathcal{R}(\boldsymbol{\phi}) = \sum_i \sum_j \ln (\sigma_{i,j}^2) .
\end{equation}
The weight $\lambda$ for balancing between the loss terms is determined empirically. Since predictive uncertainty can be regarded as inverse predictability, we derive predictability by simply nomalising the uncertainty value and compute
\begin{equation}
\phi_{i,j} = 1 - \operatorname{norm}(\sigma_{i,j}^2).
\end{equation}

\subsection{Predictability Analysis with Local Variance}
This study makes comparisons with a traditional statistical analysis for benchmarking purposes to validate that the learning-based analysis is competitive. Traditional methods often treat predictability as pattern complexity and estimate it by computing local variance of pixel intensities~\cite{2009_4811982, 2011_5762603, Peng:2012aa, Cao:2019aa}. For each query pixel, the model computes its local variance from the neighbouring context pixels. This model can be perceived as a variance filter such that the output value is the variance of the $4$-connected neighbourhood around the corresponding input pixel:
\begin{equation}
\sigma_{i,j}^2 = \frac{1}{N}\sum_{k=1}^N (\operatorname{nhood}_{k}(x_{i.j}) - \mu_{i,j})^2 ,
\end{equation}
where $\operatorname{nhood}_{k}(x_{i.j})$ denotes an element of the pixel neighbourhood, $N$ the number of neighbours, and $\mu_{i,j}$ the mean of the neighbours. The number of neighbouring pixels is $4$ in accordance with the previously defined pixel connectivity, but can be less than $4$ for pixels on the borders if border padding is not applied. A small local variance indicates a low pattern complexity and vice versa.

%%% ----- SECTION ----- %%%
%%% ----- SECTION ----- %%%
%%% ----- SECTION ----- %%%
\section{Experiments}
\label{sec:exp}
The objective of our experiments is to identify the improvements delivered by predictability analysis for several aspects of reversible steganography. We carry out an ablation study to assess the contribution of the learning-based methods benchmarked against a variance-based method. We begin by describing the datasets and benchmarks. Then, we visualise the results produced by different methods. A comparison of the classification accuracy among different methods is provided. Furthermore, as the primary concern for the residual modulation algorithm, the residual distribution is closely examined with different statistical measurements for concentration. Finally, we evaluate steganographic rate\textendash distortion performance to confirm the improvements.

\subsection{Datasets and Benchmarks}
The neural networks were trained and tested on the BOSSbase dataset~\cite{2011_BOSSbase}. This dataset was developed for an academic competition regarding digital steganography and has a collection of 10,000 greyscale photographs covering a wide diversity of subjects. The training/test split ratio was 80/20. The inference set was composed of standard test images from the USC-SIPI dataset~\cite{2006_USC_SIPI}. A comparative study was carried out to assess the supervised learning (SL) and unsupervised learning (UL) methods, benchmarked against a local variance (LV) method~\cite{2009_4811982, 2011_5762603, Peng:2012aa, Cao:2019aa}. Apart from the RDN, the prior art predictive neural networks (MS-CNN~\cite{Hu:2021aa} and MemNet~\cite{Chang:2021aa}) were also used as baseline models to evaluate the rate\textendash distortion performance. In all experiments, the steganographic parameter $\alpha$ was set to $2$.

%%% Raw %%%
%%% ----- FIGURE ----- %%%
\begin{figure}[t!] % for sub figures over two columns in
% -----------------------------------
\centering
% Raw Output
% GT Maps
\subfloat[House]
{\includegraphics[width=0.24\columnwidth]{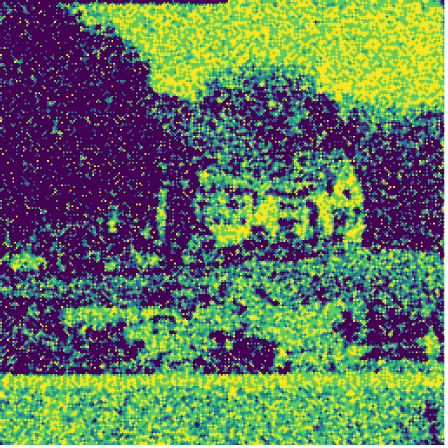}}
\hfil
\subfloat[Lake]
{\includegraphics[width=0.24\columnwidth]{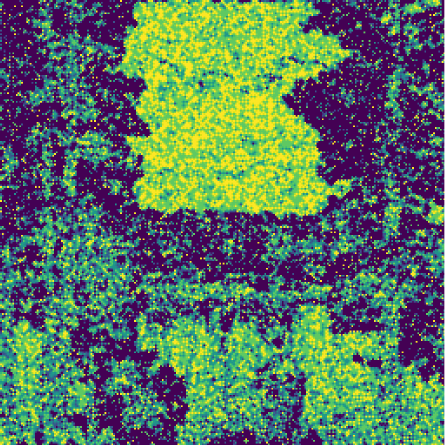}}
\hfil
\subfloat[Lena]
{\includegraphics[width=0.24\columnwidth]{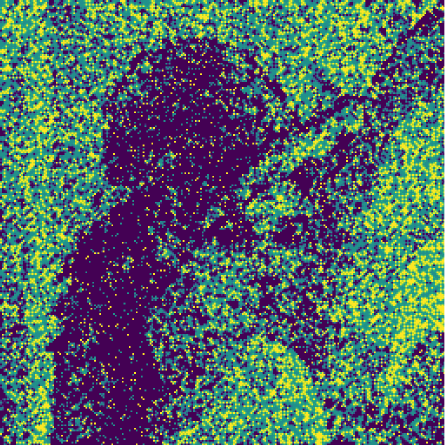}}
\hfil
\subfloat[Mandrill]
{\includegraphics[width=0.24\columnwidth]{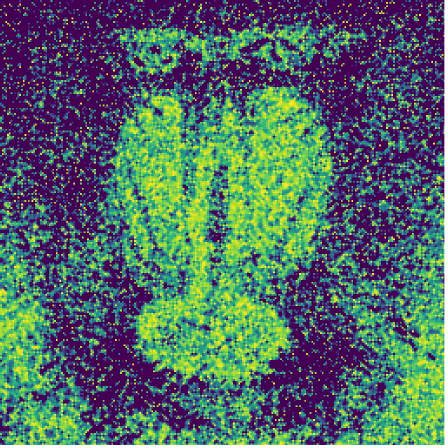}}
\\
%%%% SL
\subfloat[SL]
{\includegraphics[width=0.24\columnwidth]{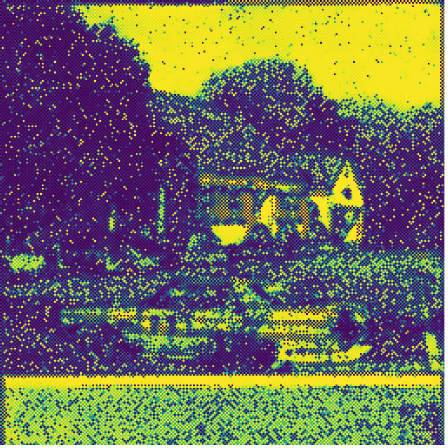}}
\hfil
\subfloat[SL]
{\includegraphics[width=0.24\columnwidth]{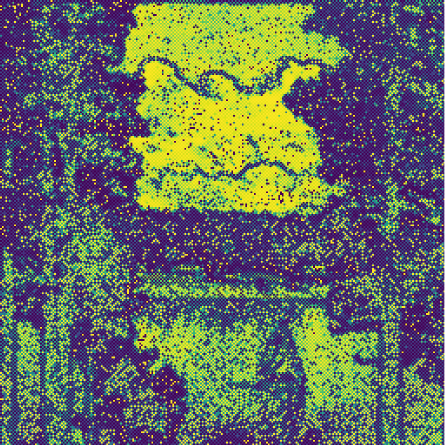}}
\hfil
\subfloat[SL]
{\includegraphics[width=0.24\columnwidth]{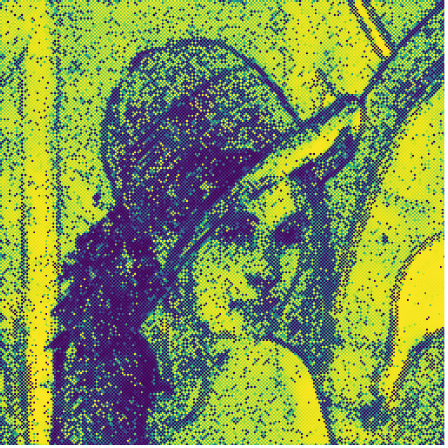}}
\hfil
\subfloat[SL]
{\includegraphics[width=0.24\columnwidth]{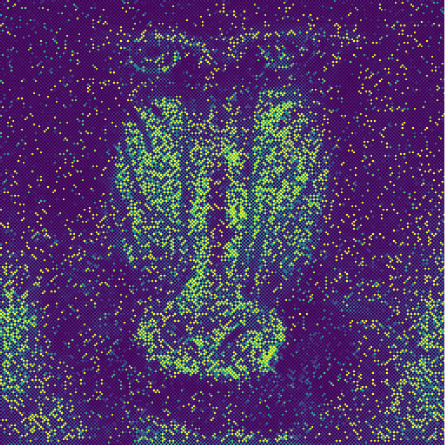}}
\\%%%% UL
\subfloat[UL]
{\includegraphics[width=0.24\columnwidth]{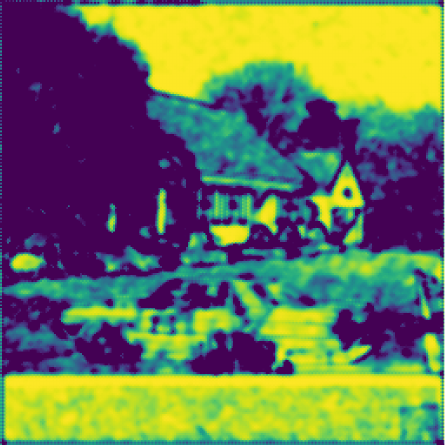}}
\hfil
\subfloat[UL]
{\includegraphics[width=0.24\columnwidth]{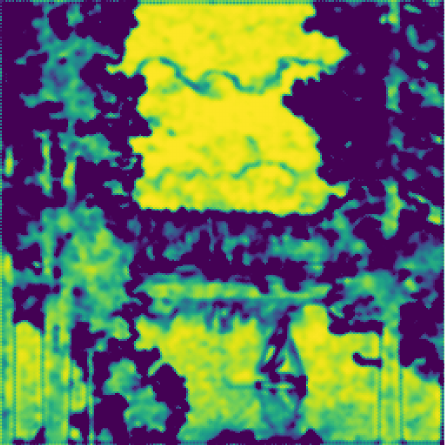}}
\hfil
\subfloat[UL]
{\includegraphics[width=0.24\columnwidth]{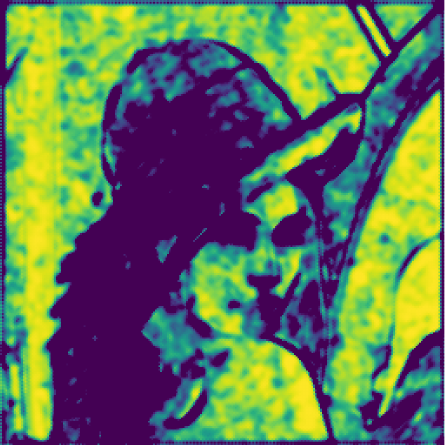}}
\hfil
\subfloat[UL]
{\includegraphics[width=0.24\columnwidth]{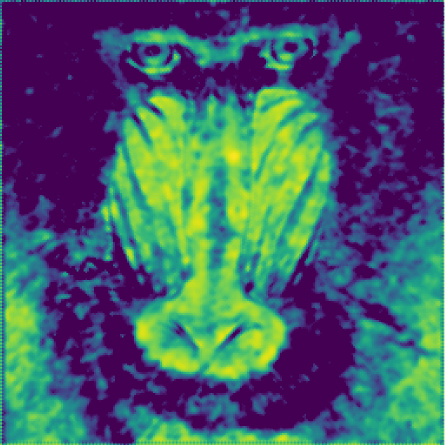}}
\\%%%% LV
\subfloat[LV]
{\includegraphics[width=0.24\columnwidth]{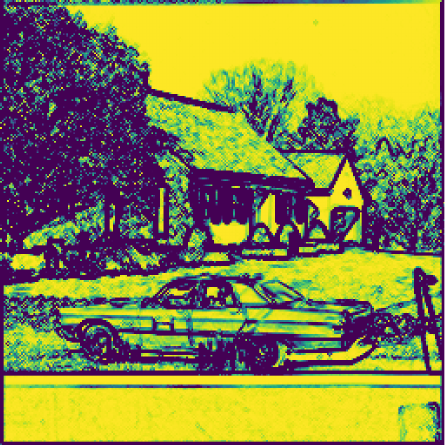}}
\hfil
\subfloat[LV]
{\includegraphics[width=0.24\columnwidth]{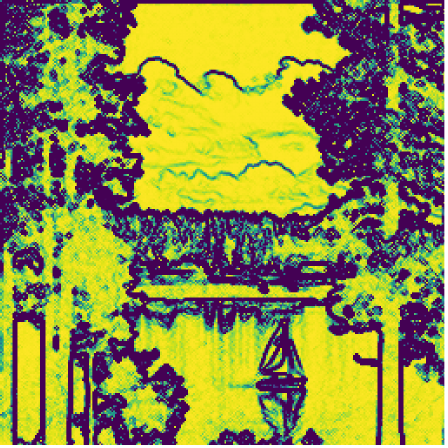}}
\hfil
\subfloat[LV]
{\includegraphics[width=0.24\columnwidth]{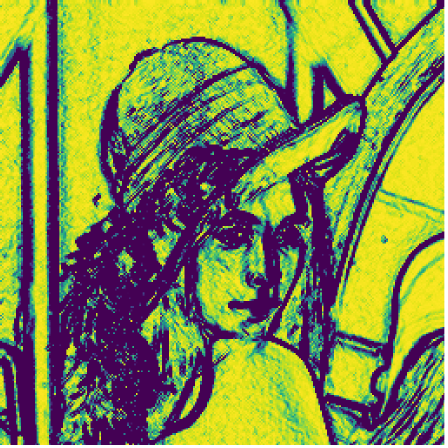}}
\hfil
\subfloat[LV]
{\includegraphics[width=0.24\columnwidth]{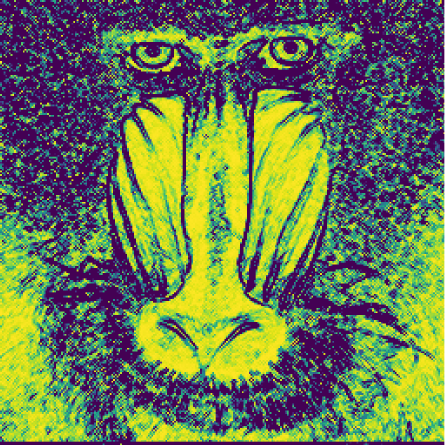}}

\caption{Visual results for residual maps (first row) and raw predictability maps of supervised learning (second row), unsupervised learning (third row) and local variance (fourth row).}
\label{fig:vis_raw}
\end{figure}
%%% ----- END OF FIGURE ----- %%%

%%% Binary %%%
%%% ----- FIGURE ----- %%%
\begin{figure}[t!] % for sub figures over two columns in
% -----------------------------------
\centering
% Raw Output
% GT Maps (query)
\subfloat[$46.3440\%$]
{\includegraphics[width=0.24\columnwidth]{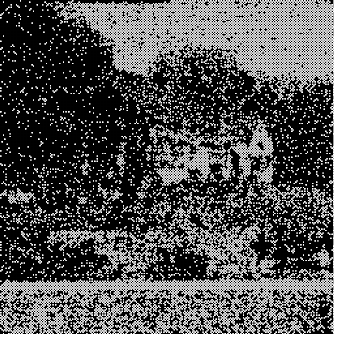}}
\hfil
\subfloat[$43.7592\%$]
{\includegraphics[width=0.24\columnwidth]{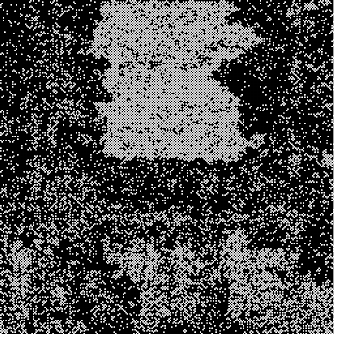}}
\hfil
\subfloat[$59.0515\%$]
{\includegraphics[width=0.24\columnwidth]{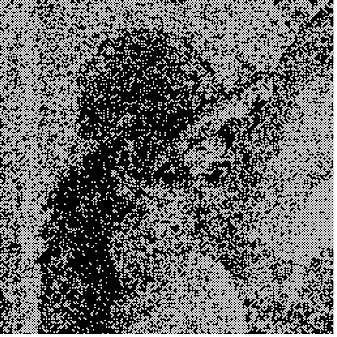}}
\hfil
\subfloat[$18.7958\%$]
{\includegraphics[width=0.24\columnwidth]{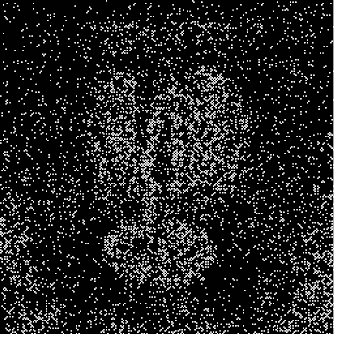}}
\\
\subfloat[$0.7661$]
{\includegraphics[width=0.24\columnwidth]{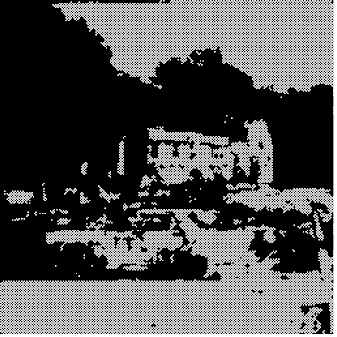}}
\hfil
\subfloat[$0.7037$]
{\includegraphics[width=0.24\columnwidth]{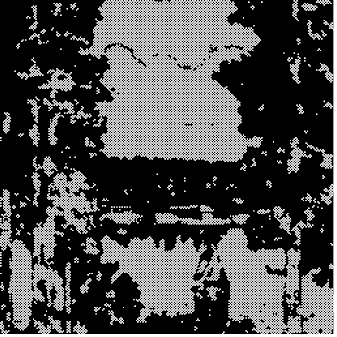}}
\hfil
\subfloat[$0.7609$]
{\includegraphics[width=0.24\columnwidth]{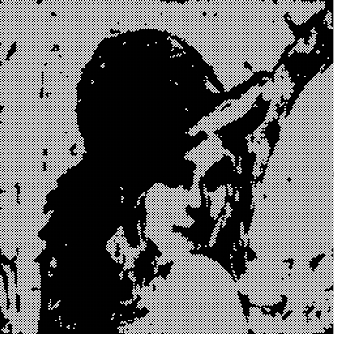}}
\hfil
\subfloat[$0.3885$]
{\includegraphics[width=0.24\columnwidth]{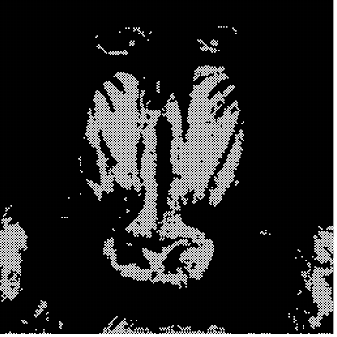}}
\\%%%% UL
\subfloat[$0.7436$]
{\includegraphics[width=0.24\columnwidth]{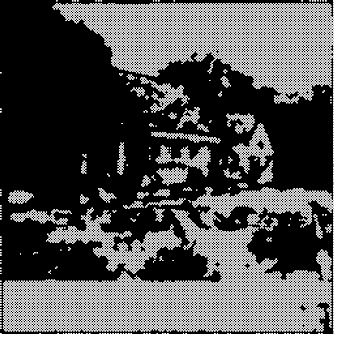}}
\hfil
\subfloat[$0.6967$]
{\includegraphics[width=0.24\columnwidth]{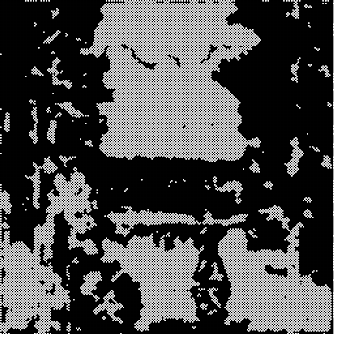}}
\hfil
\subfloat[$0.7452$]
{\includegraphics[width=0.24\columnwidth]{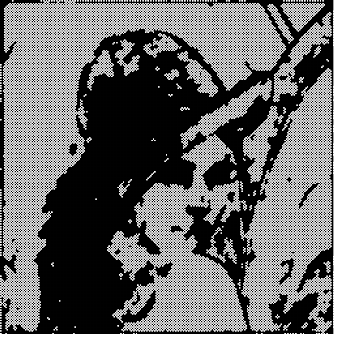}}
\hfil
\subfloat[$0.3877$]
{\includegraphics[width=0.24\columnwidth]{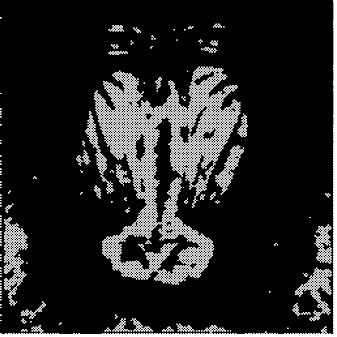}}
\\%%%% LV
\subfloat[$0.6910$]
{\includegraphics[width=0.24\columnwidth]{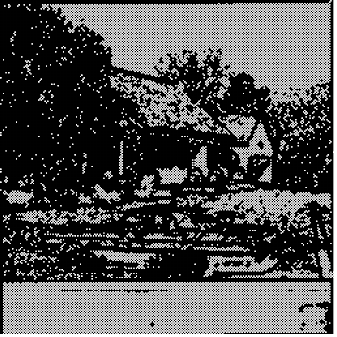}}
\hfil
\subfloat[$0.6742$]
{\includegraphics[width=0.24\columnwidth]{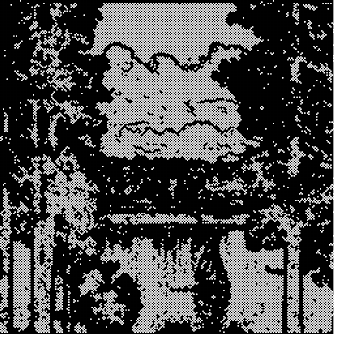}}
\hfil
\subfloat[$0.7080$]
{\includegraphics[width=0.24\columnwidth]{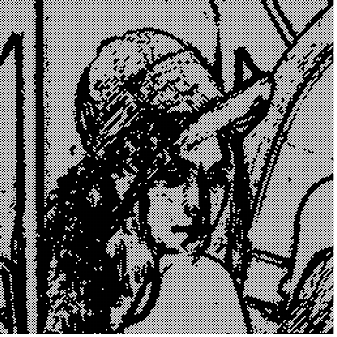}}
\hfil
\subfloat[$0.3569$]
{\includegraphics[width=0.24\columnwidth]{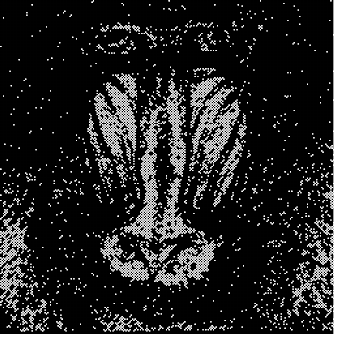}}

\caption{Visual results for carrier pixels (first row) and binarised maps of supervised learning (second row), unsupervised learning (third row) and local variance (fourth row). Numerical data in the first row represents the percentage of carrier pixels over query pixels and that in the rest rows represents precision/recall.}
\label{fig:vis_binary}
\end{figure}
%%% ----- END OF FIGURE ----- %%%

%%% ROC %%%
%%% ----- FIGURE ----- %%%
\begin{figure}[t!] % for sub figures over two columns in
% -----------------------------------
%\centering
%\begin{minipage}{.45\textwidth}
\centering
% ROC
\subfloat[House]
{\includegraphics[width=0.48\columnwidth]{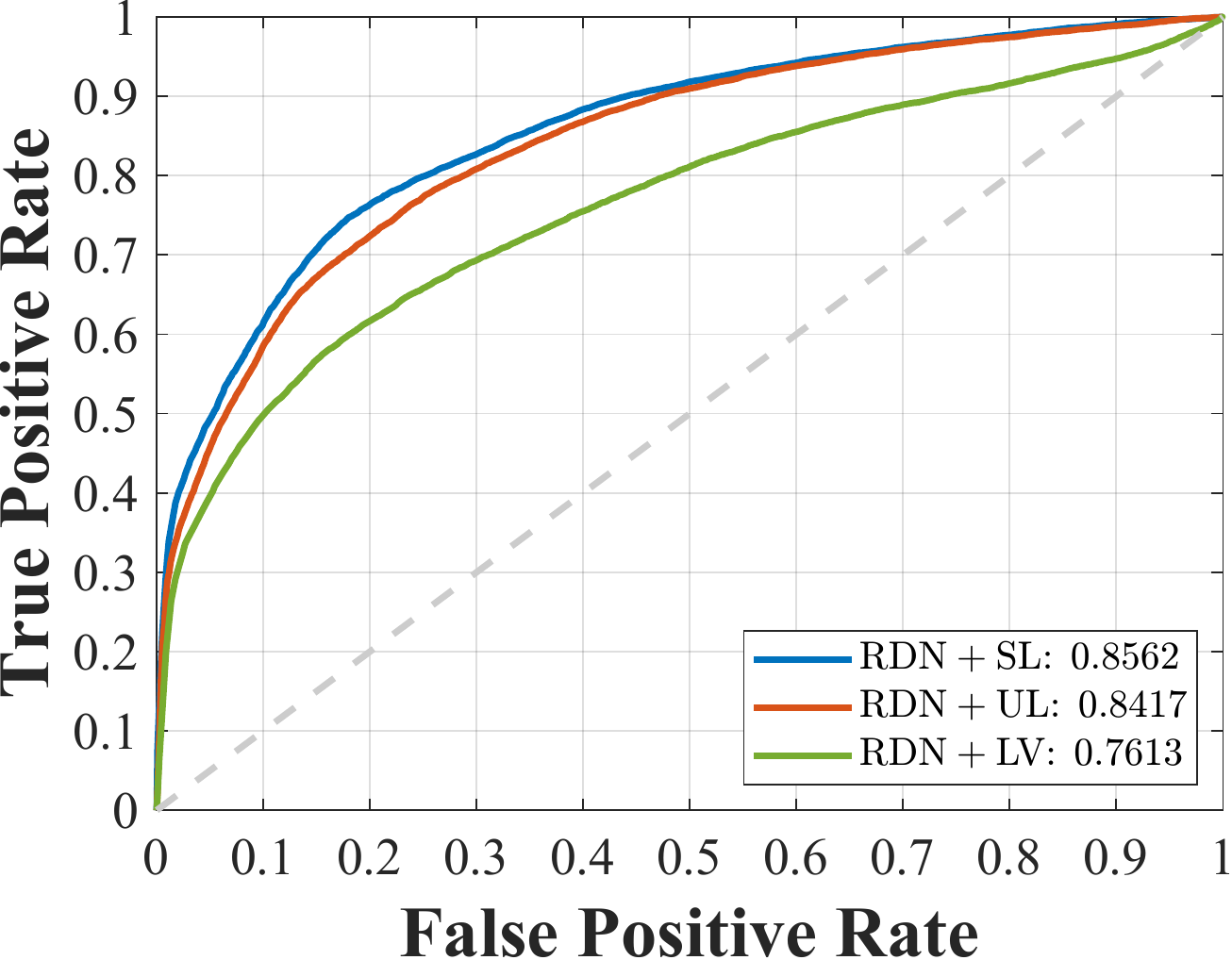}}
\hfil
\subfloat[Lake]
{\includegraphics[width=0.48\columnwidth]{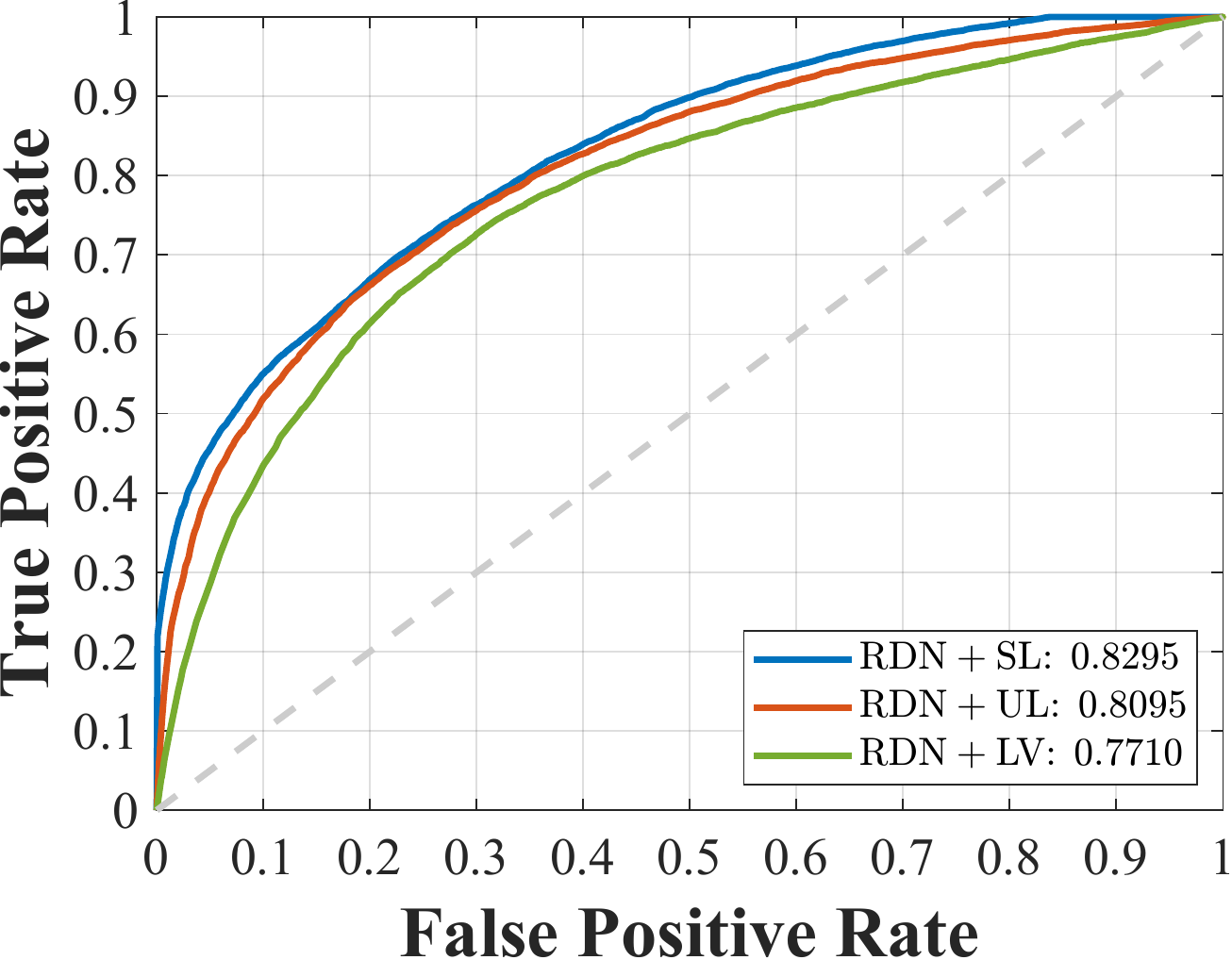}}
\hfil
\subfloat[Lena]
{\includegraphics[width=0.48\columnwidth]{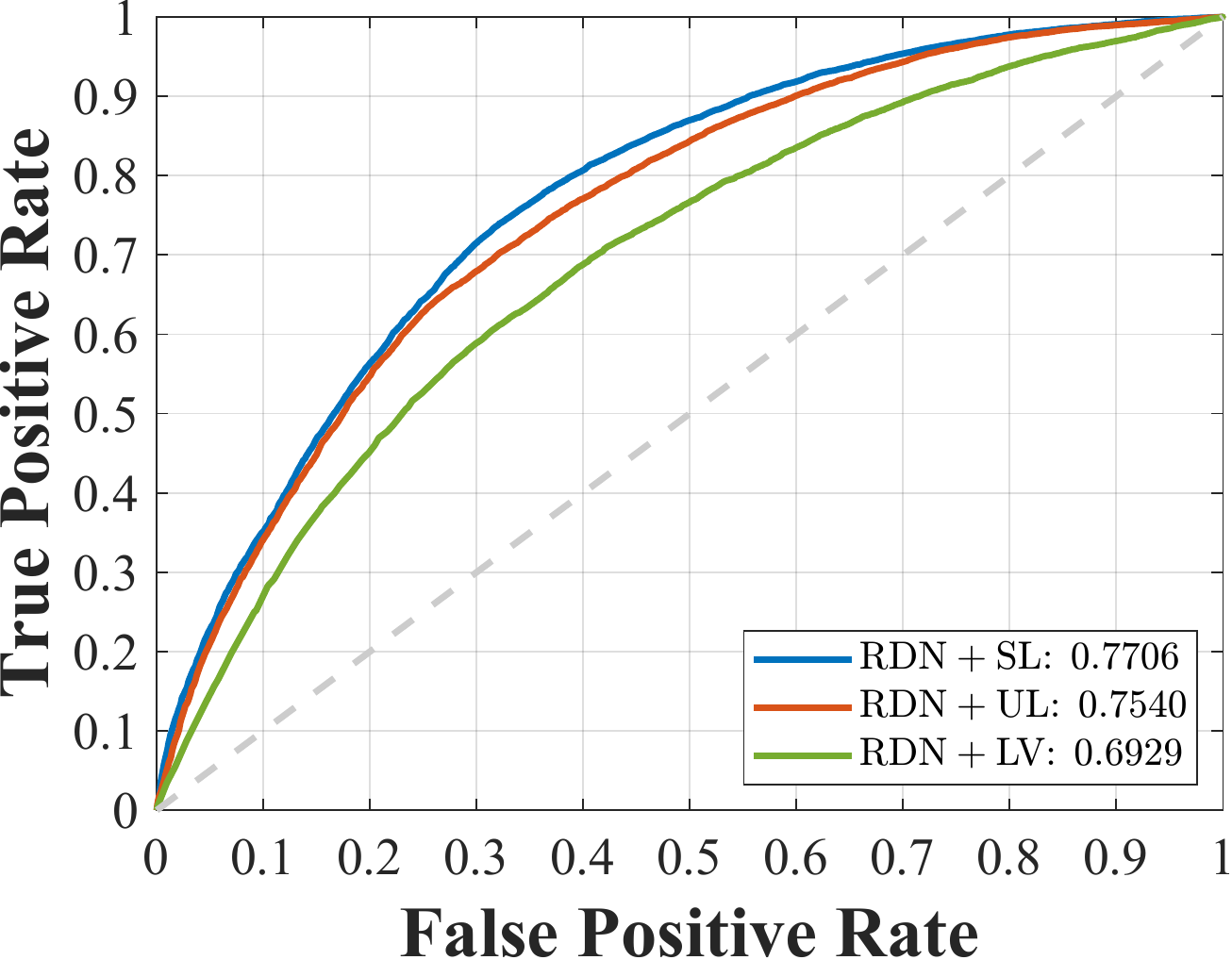}}
\hfil
\subfloat[Mandrill]
{\includegraphics[width=0.48\columnwidth]{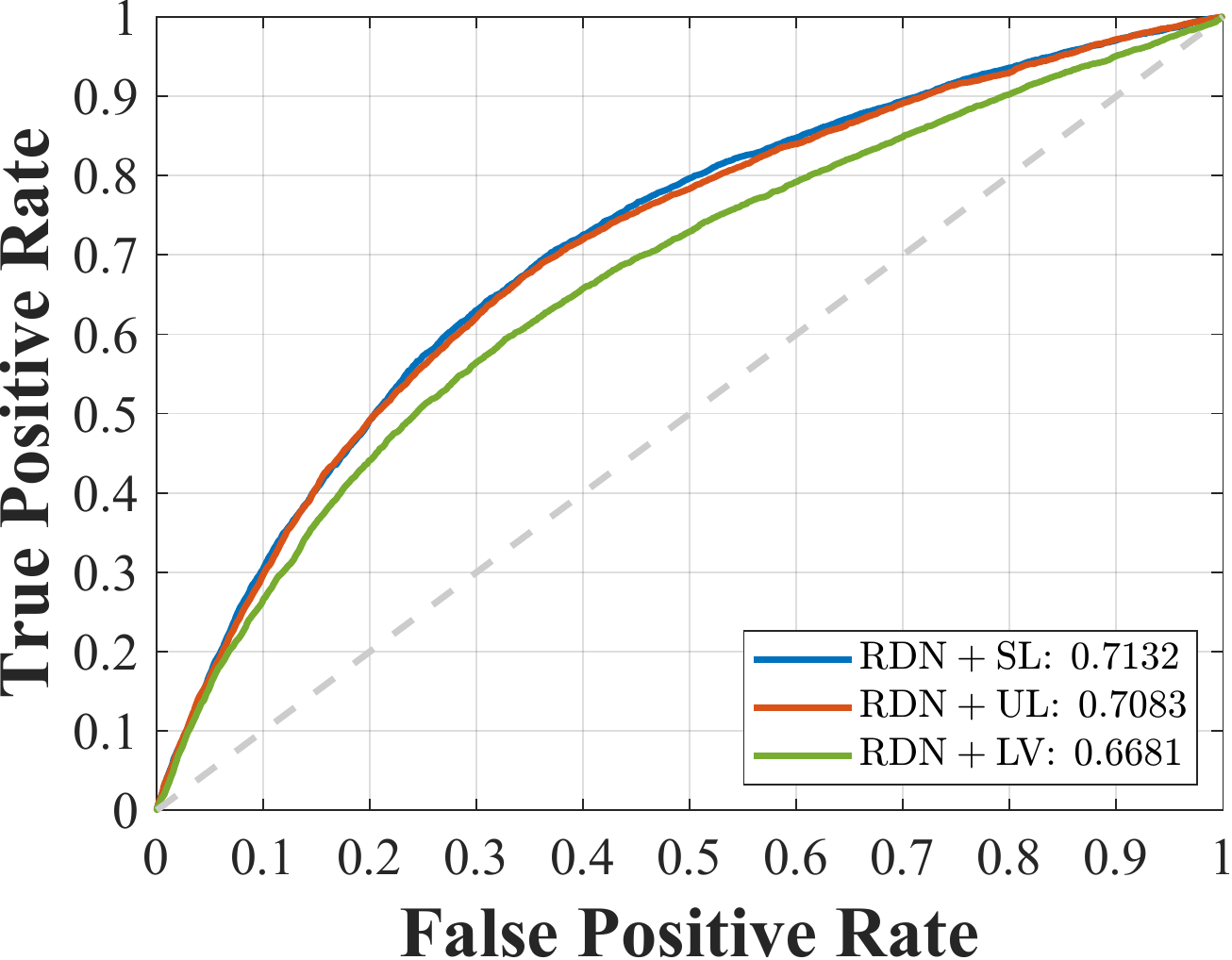}}

\caption{ROC curves with numerical data denoting AUC.}
\label{fig:exp_roc}
%\end{figure}
%\end{minipage}
%%% ----- END OF FIGURE ----- %%%
%\begin{minipage}{.5\textwidth}
%\centering
%%% Box Plot %%%
%%% ----- FIGURE ----- %%%
%\begin{figure}[t]
\vspace{2em}
\centerline{\includegraphics[width=0.98\columnwidth]{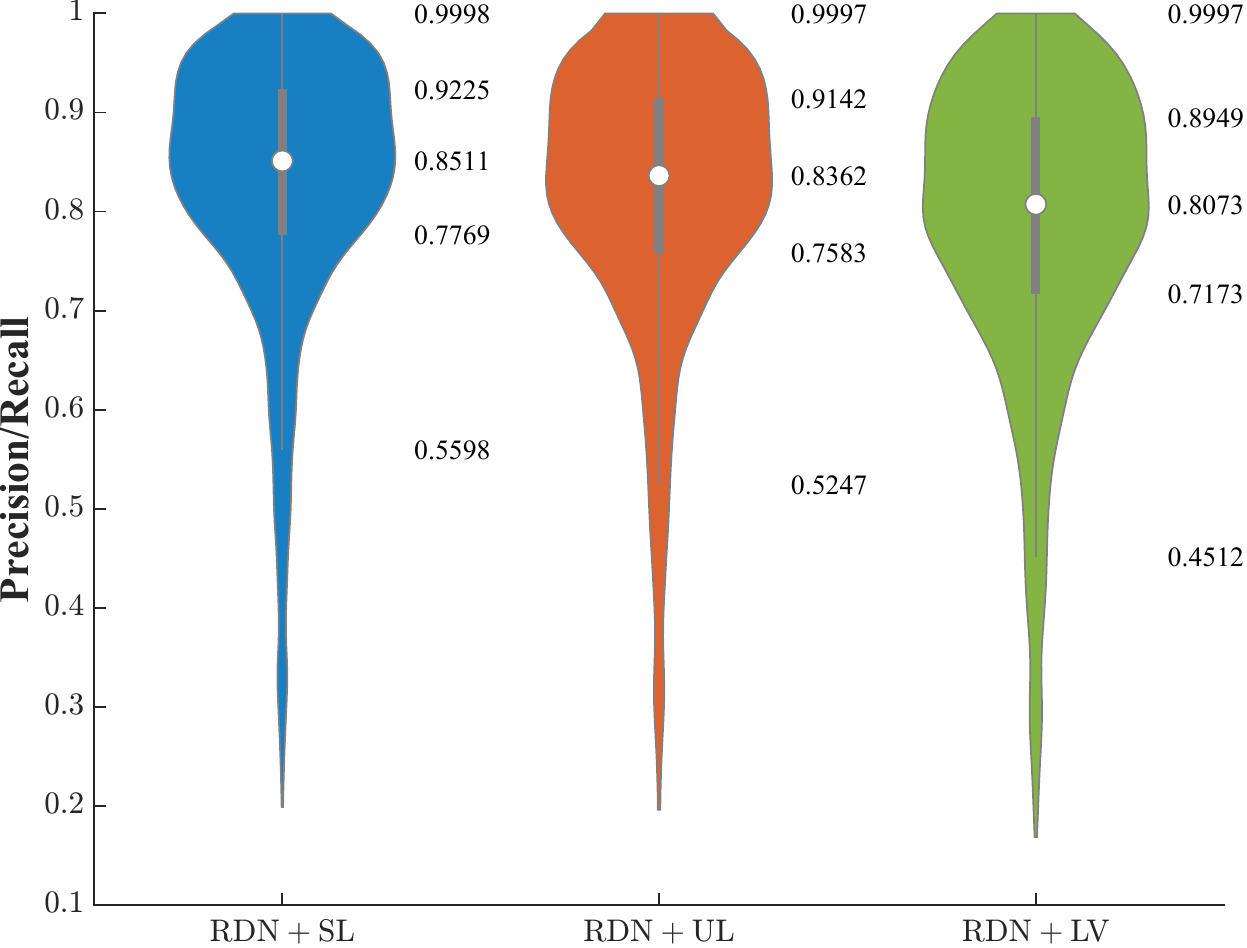}}
\caption{Prevision/recall evaluated on test set.}
\label{fig:exp_accuracy}
%\end{minipage}
\end{figure}
%%% ----- END OF FIGURE ----- %%%

%%% PDF %%%
%%% ----- FIGURE ----- %%%
\begin{figure}[t!] % for sub figures over two columns in
% -----------------------------------
%\centering
%\begin{minipage}{.5\textwidth}
\centering
% PDF
\subfloat[House]
{\includegraphics[width=0.48\columnwidth]{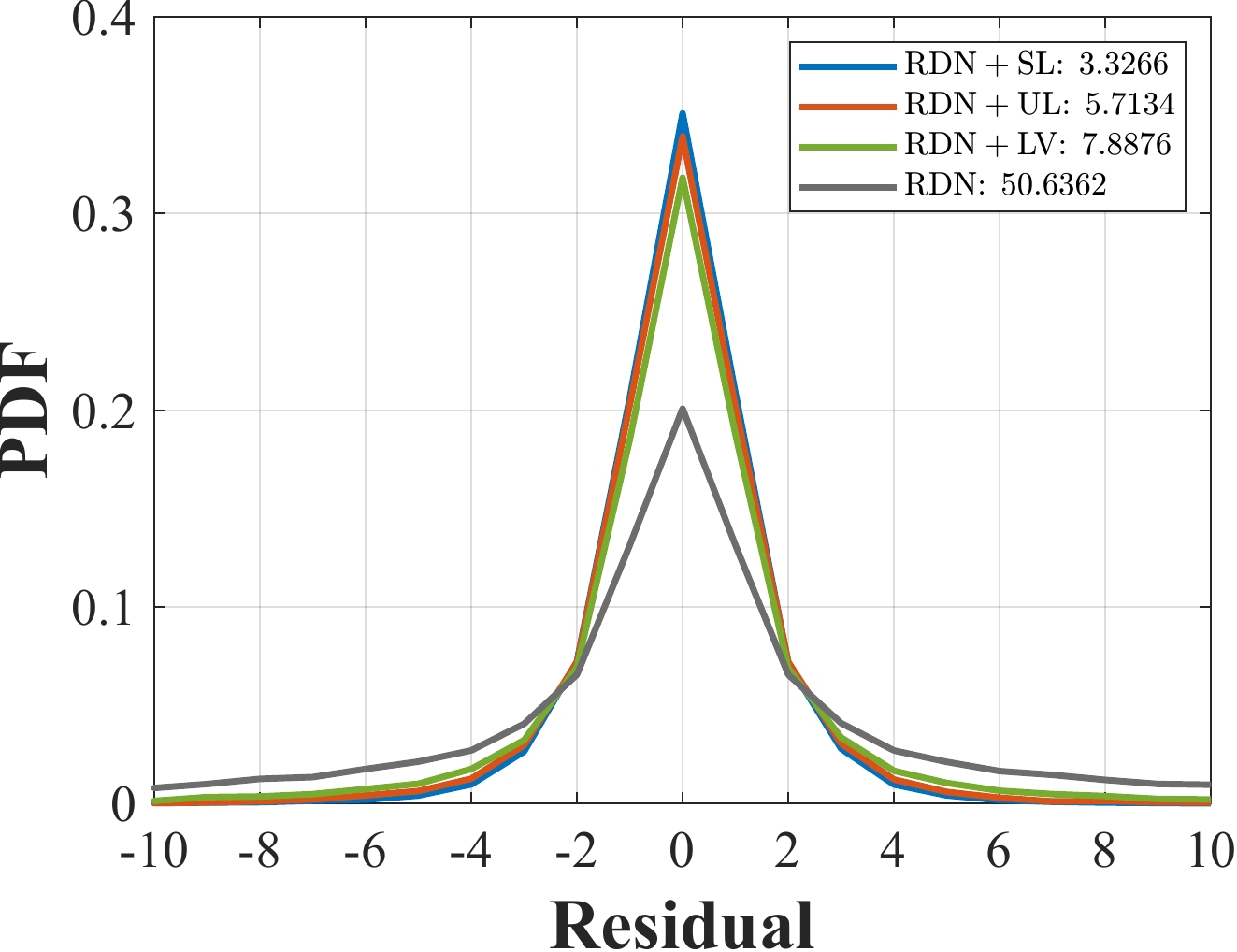}}
\hfil
\subfloat[Lake]
{\includegraphics[width=0.48\columnwidth]{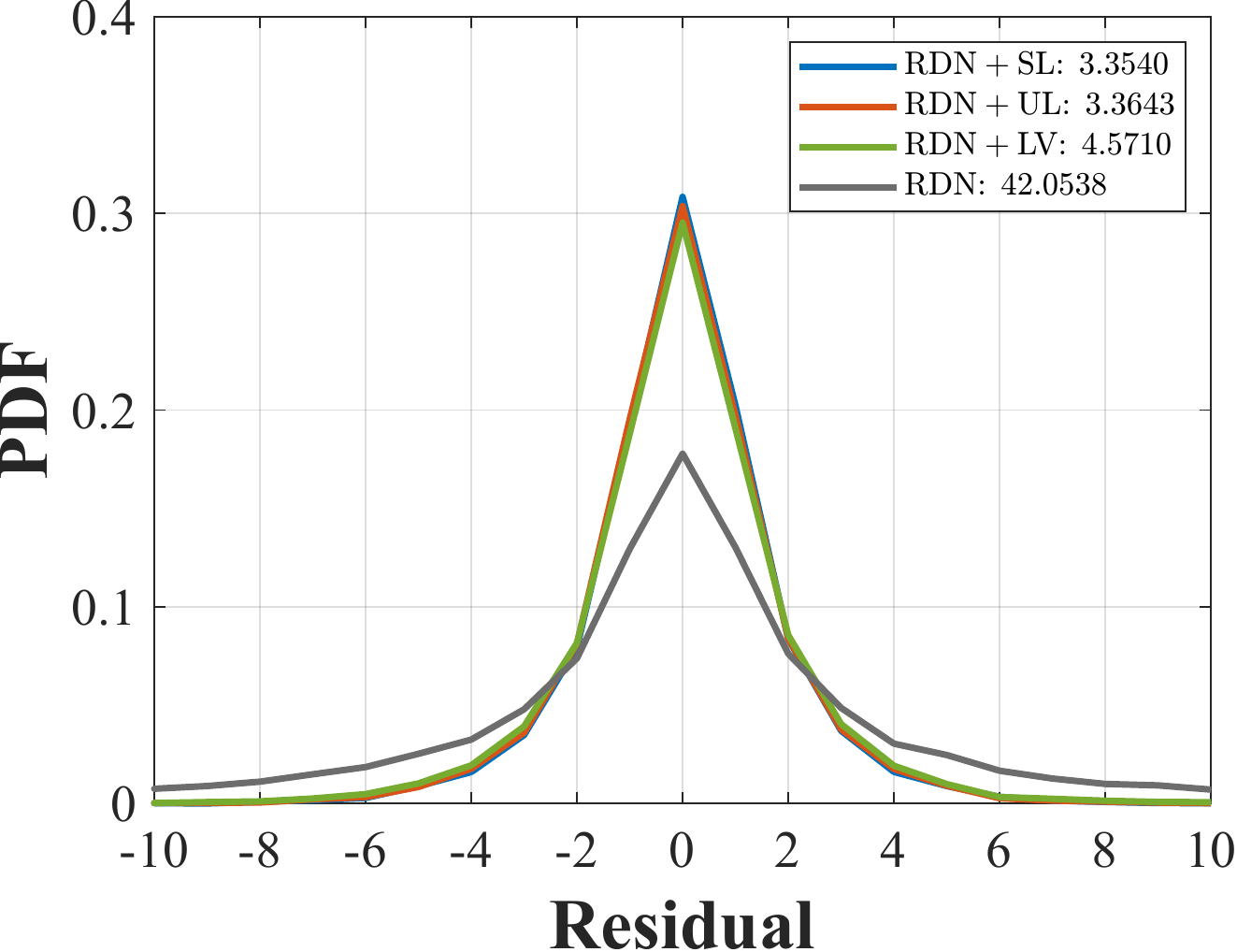}}
\hfil
\subfloat[Lena]
{\includegraphics[width=0.48\columnwidth]{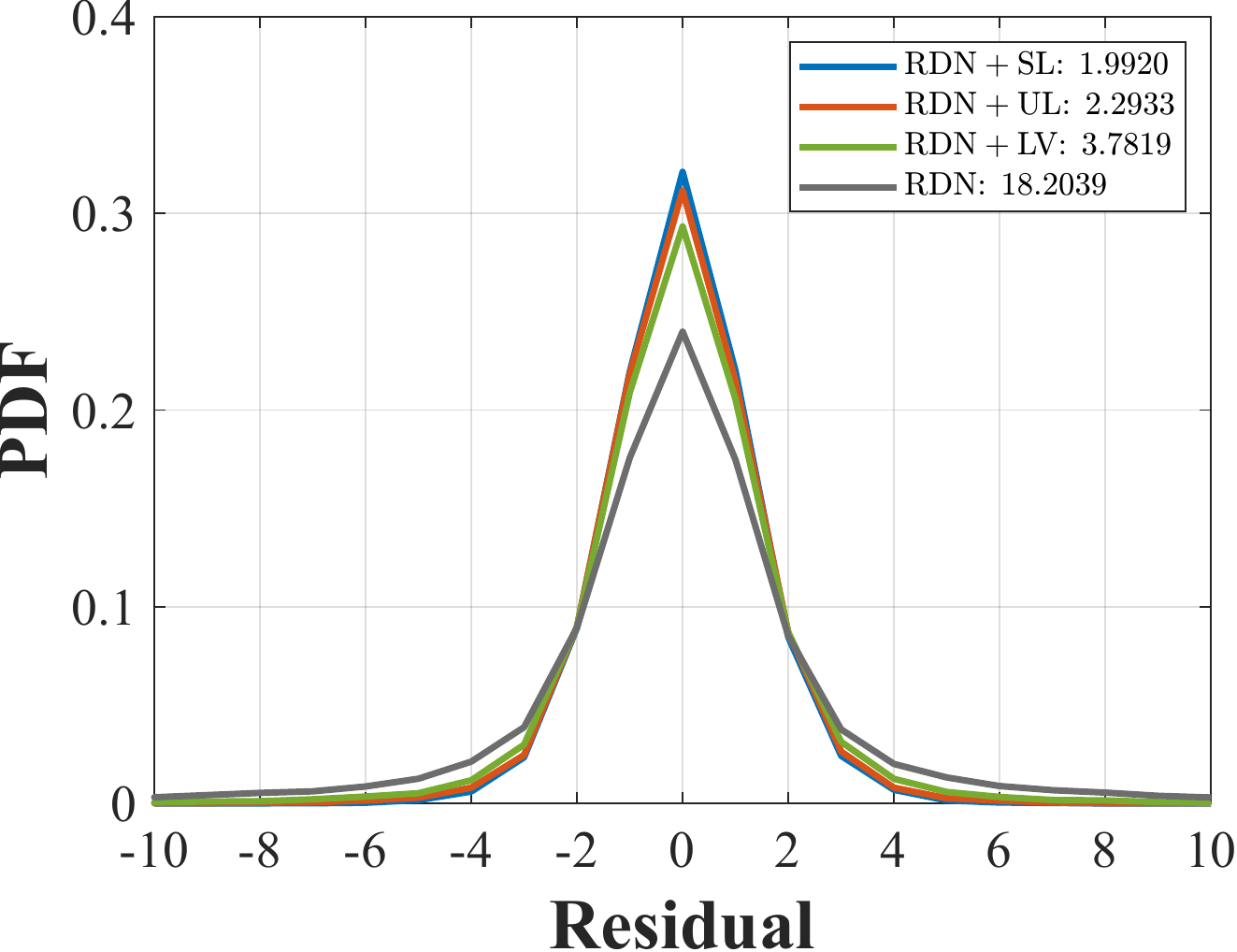}}
\hfil
\subfloat[Mandrill]
{\includegraphics[width=0.48\columnwidth]{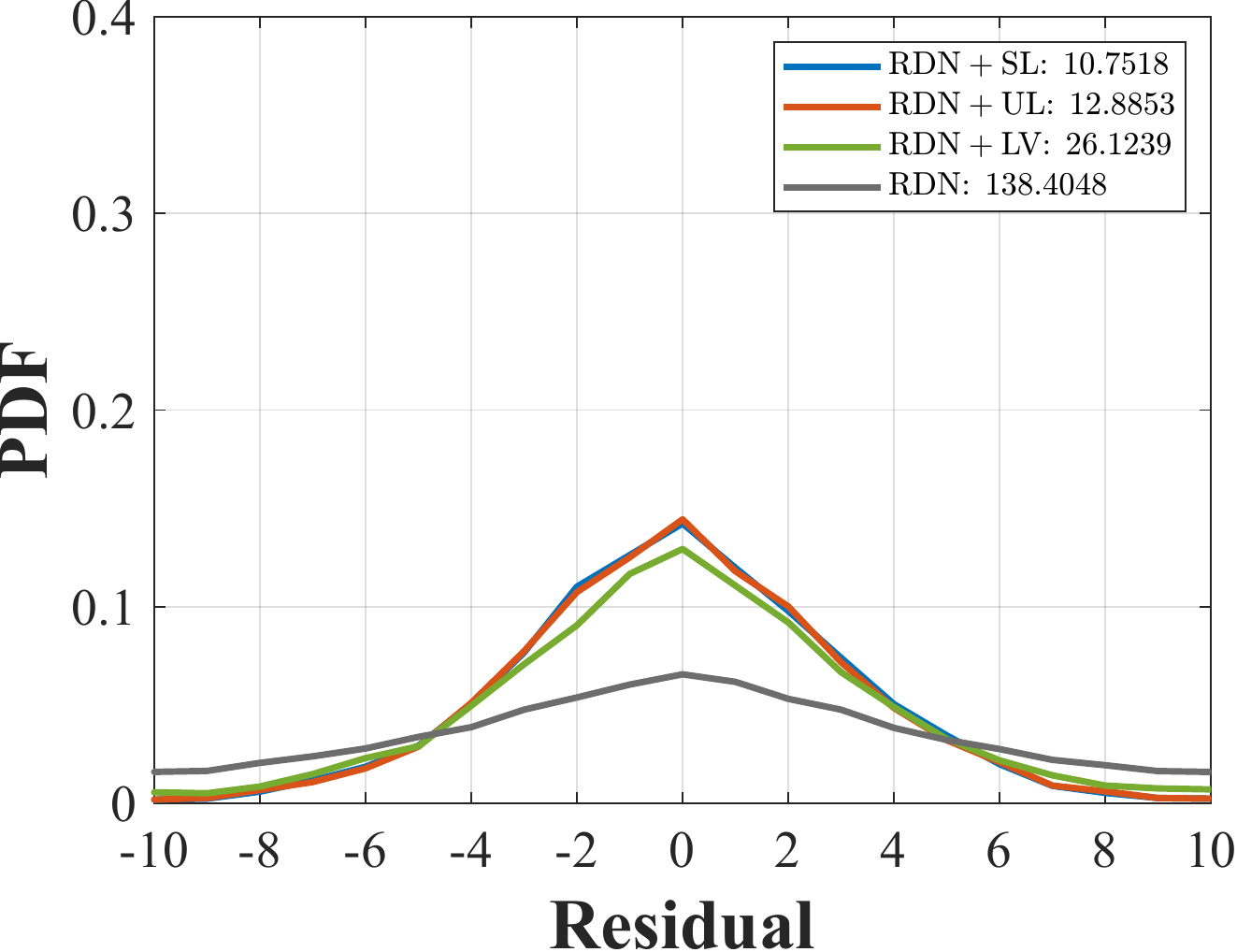}}
\caption{PDF of residual distribution with numerical data denoting variance.}
\label{fig:exp_pdf}
%\end{figure}
%\end{minipage}
%%% ----- END OF FIGURE ----- %%%
%\begin{minipage}{.49\textwidth}
%\centering
%%% Box Plot %%%
%%% ----- FIGURE ----- %%%
%\begin{figure}[t]
\vspace{2em}
\centerline{\includegraphics[width=0.98\columnwidth]{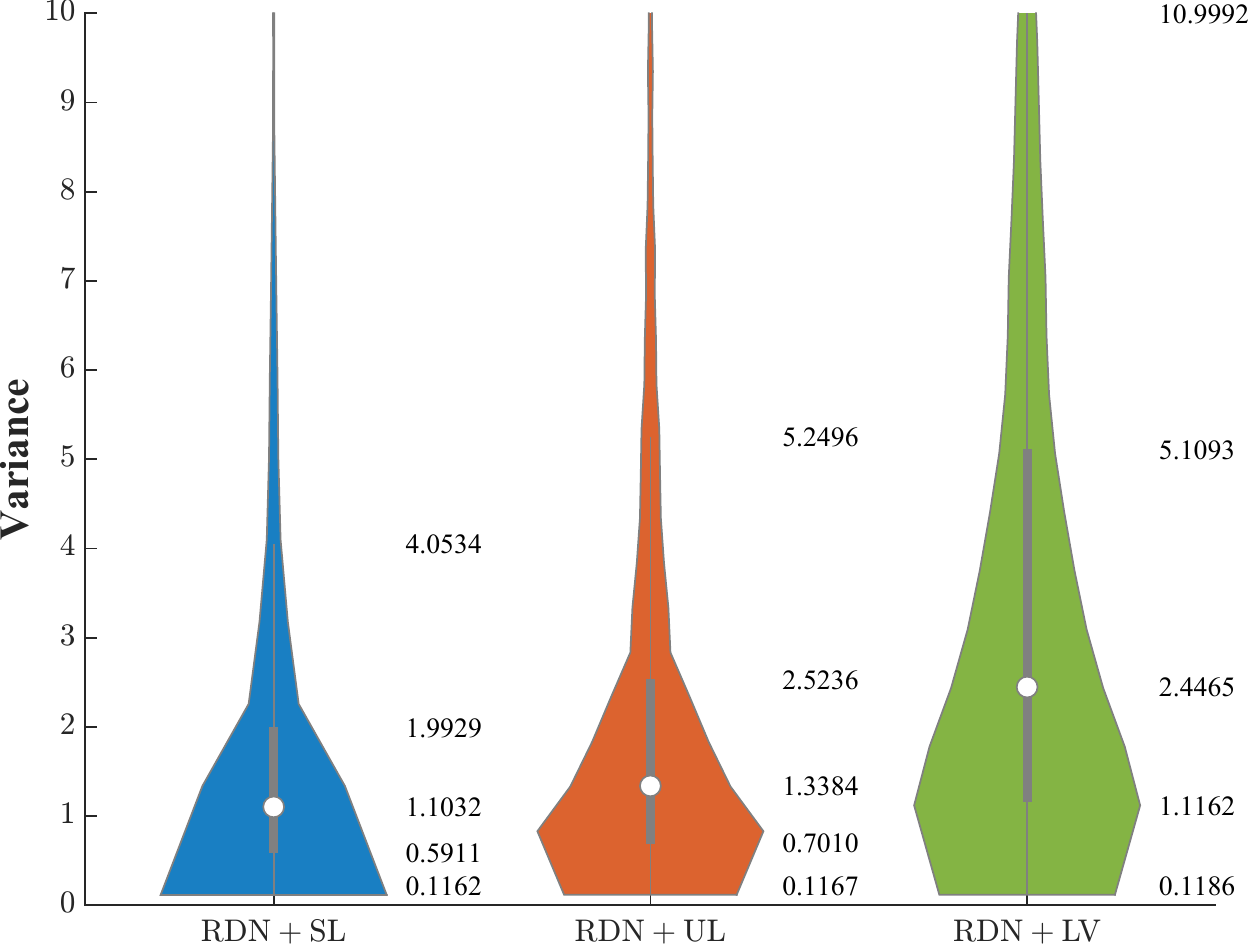}}
\caption{Variance evaluated on test set.}
\label{fig:exp_var}
%\end{minipage}
\end{figure}
%%% ----- END OF FIGURE ----- %%%

\subsection{Visual Comparison}
The objective of predictability analysis is identifying predictable pixels such that the identified pixels match the carrier pixels. The optimal solution is the binary ground-truth map (i.e. the binarised residual map) constructed for supervised learning. However, the maps produced by supervised learning, unsupervised learning and local variance are composed of continuous numeric values. Figure~\ref{fig:vis_raw} displays the residual maps and raw predictability maps of supervised learning, unsupervised learning and local variance. The lighter colour indicates a smaller residual, higher predictability, lower uncertainty or lower variance. For a fair comparison, we binarise the raw outputs in such a way that each binary map consists of the same proportion of query pixels determined as predictable. We set the proportion as the number of carrier pixels (in the ground-truth map) divided by the total number of query pixels (i.e. approximately a half of the image). Figure~\ref{fig:vis_binary} visualises the carrier pixels (by assigning zeros to the context pixels in the ground-truth map) and the binarised maps. It can be seen that the predictable pixels are mostly distributed in smooth areas, conforming to the basic statistical properties. We measure classification performance by precision and recall. The former indicates the fraction of the actual carrier pixels among the identified predictable pixels, while the latter represents the fraction of retrieved carrier pixels. In our setting, the number of pixels identified as predictable equals the number of carrier pixels and thus precision equals recall. It is shown that classificaion performance in descending order are supervised learning, unsupervised learning and local variance, which suggests that the learning-based analysers are more capable of capturing true predictability associated with both the image content and the predictor in use.

\subsection{Classification Accuracy}
We use the receiver operating characteristic (ROC) curve to further examine the diagnostic ability of different analysers on the inference set, as plotted in Figure~\ref{fig:exp_roc}. The curves show the true positive rate against the false-positive rate at various binarisation threshold values. Each image is considered as a classification dataset and each pixel as a sample to be classified. The greater the deviation from the diagonal line (line of no-discrimination), the better the classification performance. The performance can be evaluated numerically by the area under the curve (AUC). Both learning-based analysers outperform the variance-based method, and the supervised one outdoes the unsupervised one slightly. Figure~\ref{fig:exp_accuracy} depicts precision/recall in a large-scale assessment on the test set using a five-number summary: the minimum, the first quartile, the median, the third quartile, and the maximum. The trend for the large-scale assessment is similar to that for the individual image samples.

%%% Rate_Dist %%%
%%% ----- FIGURE ----- %%%
\begin{figure*}[t!] % for sub figures over two columns in
% -----------------------------------
\centering
% Aeroplane
\subfloat[House]
{\includegraphics[width=0.9\columnwidth]{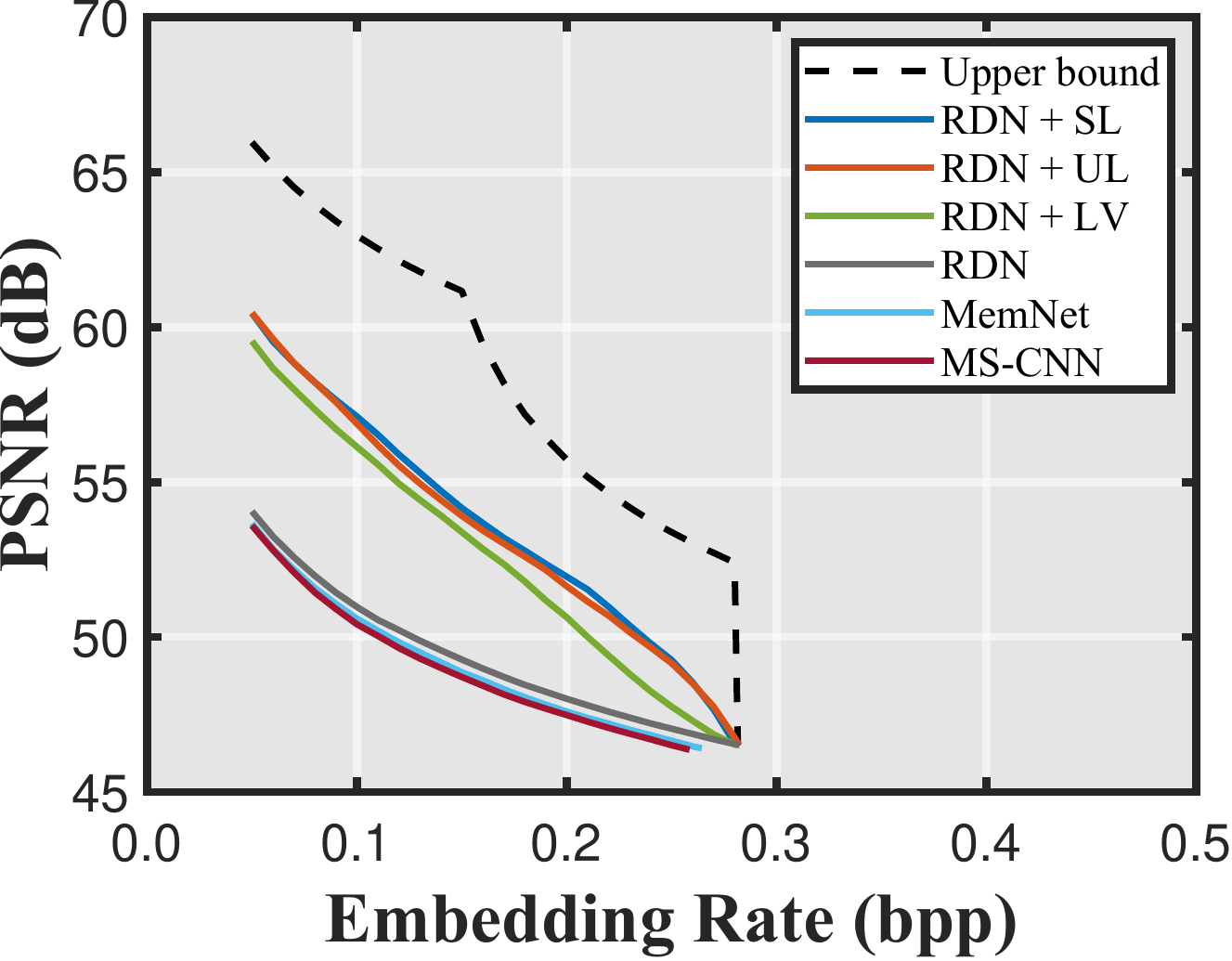}}
\hfil
\subfloat[Lake]
{\includegraphics[width=0.9\columnwidth]{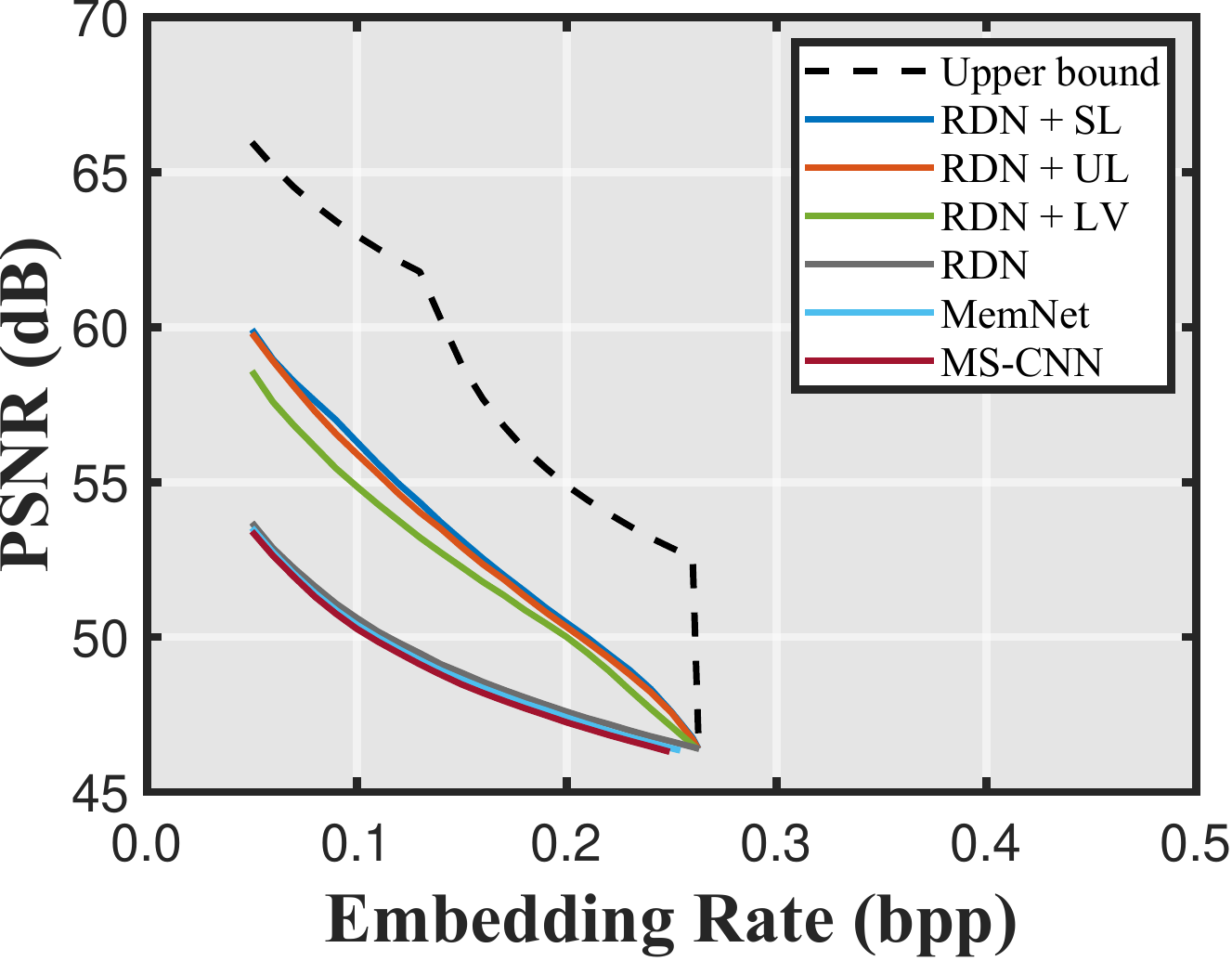}}
\\
\subfloat[Lena]
{\includegraphics[width=0.9\columnwidth]{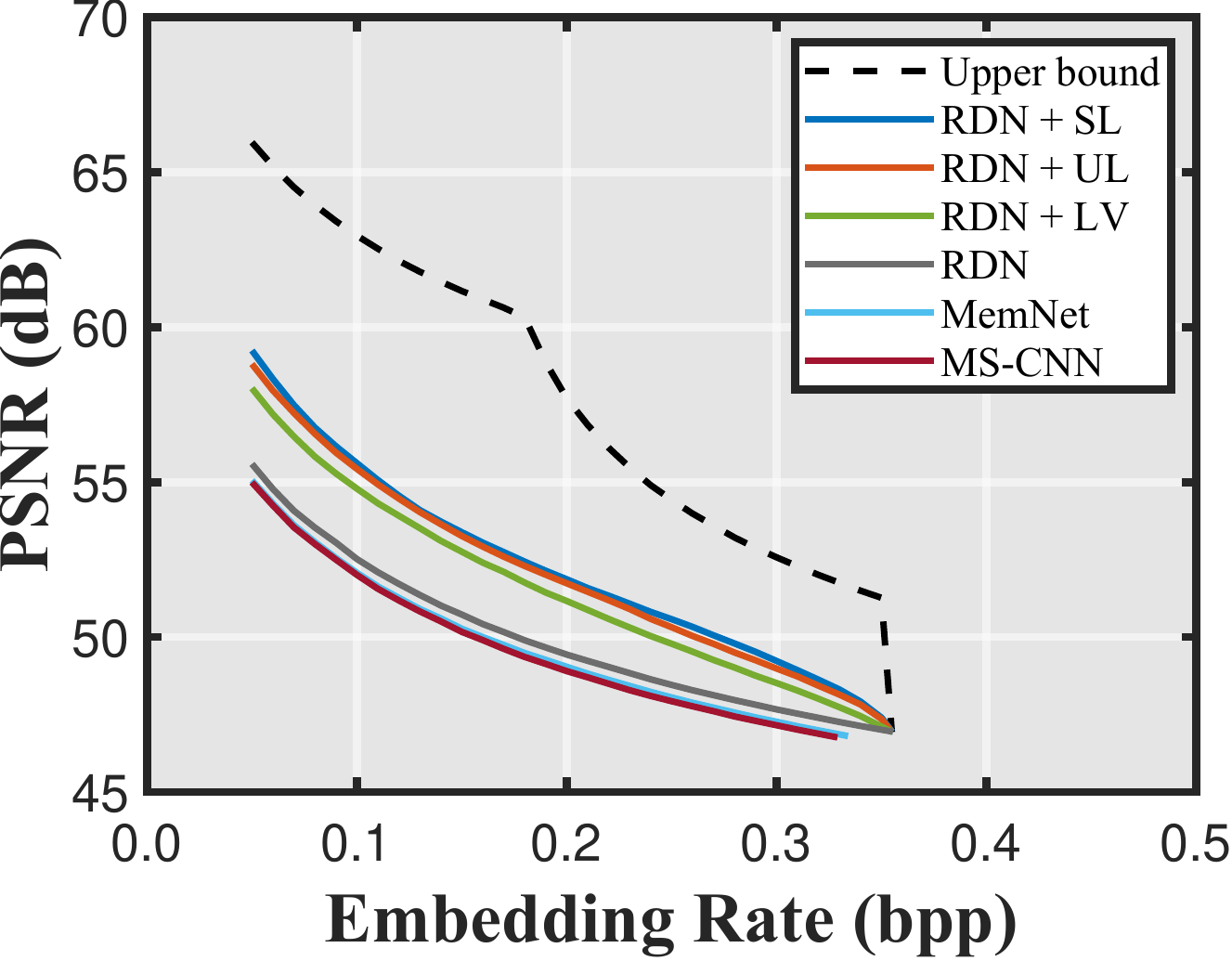}}
\hfil
\subfloat[Mandrill]
{\includegraphics[width=0.9\columnwidth]{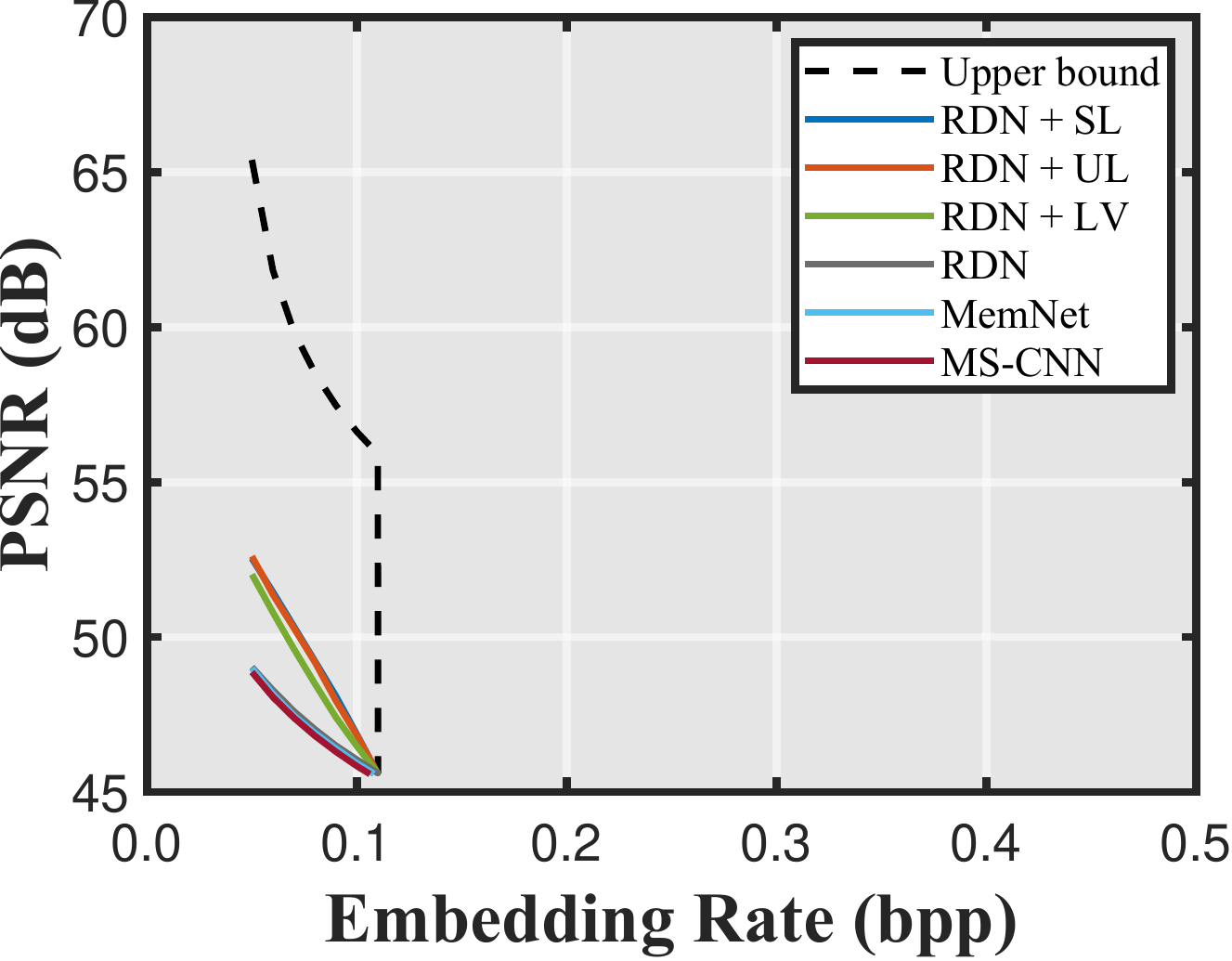}}

\caption{Steganographic rate\textendash distortion curves benchmarked against predictors with and without abstention option.}
\label{fig:exp_rate_dist}
\end{figure*}
%%% ----- END OF FIGURE ----- %%%

\subsection{Residual Distribution}
The residual modulation algorithm is designed based on the observation that residuals normally follow a Laplacian distribution and that a more accurate predictive model yields a more concentrated distribution. The residual distribution is a determining factor affecting steganographic rate\textendash distortion performance. Figure~\ref{fig:exp_pdf} depicts the residual distribution of each individual image by the probability distribution function (PDF) labelled with the distribution's variance as a measurement for concentration. A smaller variance score indicates a more concentrated residual distribution. It is observed that predictability analysers indeed lead to a more concentrated residual distribution by filtering out unpredictable pixels. Figure~\ref{fig:exp_var} shows the distribution of variance scores on the test set with a five-number summary. In this large-scale assessment, the performance in descending order are supervised learning, unsupervised learning and local variance.

\subsection{Rate\textendash Distortion Performance}
The results for steganographic rate\textendash distortion performance are shown in Figure \ref{fig:exp_rate_dist}. We evaluate steganographic distortion based on the peak signal-to-noise ratio (PSNR) (expressed in decibels; dB) and steganographic capacity based on the embedding rate (expressed in bits per pixel; bpp). The image quality decreases steadily as the embedding rate increases. A factor that affects the rate\textendash distortion performance is predictive accuracy. The RDN is considered as the state-of-the-art predictive model and it indeed achieves a better trade-off between capacity and imperceptibility than the MemNet and MS-CNN. The upper bound represents an embedding route by using the ground-truth map. It is shown that the provision of predictability analysers improves steganographic performance significantly, although there is still a performance gap from the upper bound. Overall, the RDN with a supervised analyser is ranked ahead, followed by that with an unsupervised analyser. The variance-based analyser is ranked in last place, confirming the validity of both supervised and unsupervised frameworks for predictability analysis.

%%% ----- SECTION ----- %%%
%%% ----- SECTION ----- %%%
%%% ----- SECTION ----- %%%
\section{Conclusion}
\label{sec:con}
Although reversible steganography with deep learning is still in its infancy, systems that transform some of their modules into neural networks have already been proven to outperform their original version. In this paper, we further investigate learning-based predictability analysers for identifying potential carrier pixels to improve steganographic trade-off between capacity and imperceptibility. We propose both supervised and unsupervised learning frameworks for predictability analysis. The experimental results verify the validy of both learning-based analysers with a substantial improvement over a traditional statistical analyser. The supervised and unsupervised methods could both be further improved by incorporating state-of-the-art neural network models and by rethinking the loss functions. While we incorporate neural networks into a reversible steganographic system following the modular paradigm, new development may also come from the end-to-end paradigm. We look forward to future progress on reversible steganography with deep learning.

\section*{Acknowledgments}
This work was partially supported by KAKENHI Grants (JP16H06302, JP18H04120, JP20K23355, JP21H04907 and JP21K18023) from the Japan Society for the Promotion of Science (JSPS) and CREST Grants (JPMJCR18A6 and JPMJCR20D3) from the Japan Science and Technology Agency (JST).

\section*{Statements and Declarations}
\begin{itemize}
\item Funding: 
This work was partially supported by KAKENHI Grants (JP16H06302, JP18H04120, JP20K23355, JP21H04907 and JP21K18023) from the Japan Society for the Promotion of Science (JSPS) and CREST Grants (JPMJCR18A6 and JPMJCR20D3) from the Japan Science and Technology Agency (JST).

\item Conflict of interest/Competing interests:
On behalf of all authors, the corresponding author states that there is no conflict of interest. The authors have no relevant financial or non-financial interests to disclose.

%\item Ethics approval 
%\item Consent for publication
%\item Availability of data and materials
\item Authors' contributions:
All authors contributed to the study conception and design. Material preparation, data collection and analysis were performed by Ching-Chun Chang, Xu Wang and Sisheng Chen. The first draft of the manuscript was written by Ching-Chun Chang and all authors commented on previous versions of the manuscript. All authors read and approved the final manuscript.

\end{itemize}

%\backmatter

%\IEEEtriggeratref{36}
\bibliographystyle{Transactions-Bibliography/IEEEtran}
\bibliography{./Bib/myBib_abbrv}


\end{document}